\documentclass[10pt]{article}

\usepackage[a4paper,
            bindingoffset=0.2in,
            left=1in,
            right=1in,
            top=1in,
            bottom=1in,
            footskip=.25in]{geometry}

\AtBeginDocument{%
  }

\usepackage{microtype}
\usepackage{graphicx}
\usepackage{tabularx}
\usepackage{subfigure}
\usepackage{booktabs} 
\usepackage{makecell}
\usepackage{multirow}
\usepackage{caption}
\usepackage{hyperref}

\usepackage{amsmath}  \usepackage{mathtools} \usepackage{amsthm}

\usepackage[capitalize,noabbrev]{cleveref}

\theoremstyle{plain}
\newtheorem{theorem}{Theorem}[section]
\newtheorem{proposition}[theorem]{Proposition}
\newtheorem{lemma}[theorem]{Lemma}

\newtheorem{corollary}[theorem]{Corollary}
\newtheorem{fact}[theorem]{Fact}
\theoremstyle{definition}
\newtheorem{definition}[theorem]{Definition}

\theoremstyle{remark}

\usepackage[textsize=tiny]{todonotes}

\DeclareMathOperator{\polylog}{polylog}
\DeclareMathOperator{\cost}{cost}
\DeclareMathOperator{\dist}{dist}
\DeclareMathOperator{\poly}{poly}

\newcommand{\R}{\mathbb{R}}

\newcommand{\coreset}{\Omega}
\newcommand{\calC}{\mathcal C}
\newcommand{\calD}{\mathcal D}
\newcommand{\calS}{\mathcal S}

\newcommand{\opt}{\text{OPT}}
\newcommand{\eps}{\varepsilon}

\newcommand{\fkmeans}{\textsc{Fast-kmeans++}~}

\newcommand{\lpar}{\left(}
\newcommand{\rpar}{\right)}
\newcommand{\lbra}{\left\{}
\newcommand{\rbra}{\right\}}
\newcommand{\lnor}{\left\|}
\newcommand{\rnor}{\right\|}

\newcounter{sideremark}

\usepackage[utf8]{inputenc} 
\usepackage[T1]{fontenc}    
\usepackage{hyperref}       
\usepackage{url}            
\usepackage{booktabs}       
\usepackage{amsfonts}       
\usepackage{nicefrac}       
\usepackage{microtype}      
\usepackage{xcolor}         
\usepackage{algorithm}
\usepackage[noend]{algpseudocode}

\usepackage{authblk}
\usepackage[numbers]{natbib}

\makeatletter
\g@addto@macro{\endtabular}{\rowfont{}}
\makeatother
\newcommand{\rowfonttype}{}
\newcommand{\rowfont}[1]{
\gdef\rowfonttype{#1}#1\ignorespaces%
}
\makeatother

\title{Settling Time vs. Accuracy Tradeoffs for Clustering Big Data}

\author[1]{Andrew Draganov\footnote{All authors contributed equally to this research.}}
\author[2]{David Saulpic}
\author[1]{Chris Schwiegelshohn}
\affil[1]{Aarhus University}
\affil[2]{Université Paris Cité \& CNRS}
\date{}

\begin{document}

\maketitle

\begin{abstract}
We study the theoretical and practical runtime limits of $k$-means and $k$-median clustering on large datasets. Since effectively all clustering methods are
slower than the time it takes to read the dataset, the fastest approach is to quickly compress the data and perform the clustering on the compressed
representation. Unfortunately, there is no universal best choice for compressing the number of points -- while random sampling runs in sublinear time and
coresets provide theoretical guarantees, the former does not enforce accuracy while the latter is too slow as the numbers of points and clusters grow. Indeed,
it has been conjectured that any sensitivity-based coreset construction requires super-linear time in the dataset size.

We examine this relationship by first showing that there does exist an algorithm that obtains coresets via sensitivity sampling in effectively linear time --
within log-factors of the time it takes to read the data.  Any approach that significantly improves on this must then resort to practical heuristics, leading us
to consider the spectrum of sampling strategies across both real and artificial datasets in the static and streaming settings. Through this, we show the
conditions in which coresets are necessary for preserving cluster validity as well as the settings in which faster, cruder sampling strategies are sufficient.
As a result, we provide a comprehensive theoretical and practical blueprint for effective clustering regardless of data size.  Our code is publicly available at
\textcolor{blue}{\href{https://github.com/Andrew-Draganov/Fast-Coreset-Generation}{source}} and has scripts to recreate the experiments.
\end{abstract}

\section{Introduction}

The modern data analyst has no shortage of clustering algorithms to choose from but, given the ever-increasing size of relevant datasets, many are often too slow to be practically useful.  
This is particularly relevant for big-data pipelines, where clustering algorithms are commonly used for compression. The goal is to replace a very large dataset by a smaller, more manageable one for downstream tasks, with the hope it represents the original input well. Lloyd's algorithm \cite{Lloyd82} was introduced for precisely this reason and minimizes the \emph{quantization error} -- the sum of square distance from each input point to its representative in the compressed dataset.
Arguably the most popular clustering algorithm, Lloyd's runs for multiple iterations until
convergence with every iteration requiring $O(ndk)$ time, where $n$ is the number of points, $d$ is the number of features and $k$ is the number of clusters -- or the size of the targeted compression. For such applications, the number of points can easily be hundreds of millions and, since the quality of compression increases with $k$, standard objectives can have $k$ in the thousands~\cite{large_k_example_1, large_k_example_2}. In such settings, any $O(ndk)$ algorithm is prohibitively slow. 

Examples like these have prompted the rise of big data algorithms that provide both theoretical and practical runtime improvements. The perspectives of theoretical soundness and practical efficacy are, however, often at odds with one another. On the one hand, theoretical guarantees provide assurance that the algorithm will work regardless of whatever unlucky inputs it receives. On the other hand, it may be difficult to convince oneself to implement the theoretically optimal algorithm when there are cruder methods that are faster to get running and perform well in practice.

Since datasets can be large in the number of points $n$ and/or the number of features $d$, big-data methods must mitigate the effects of both.  With respect to the feature space, the question is effectively closed as random projections are fast (running in effectively linear time), practical to implement \cite{MakarychevMR19}, and provide tight guarantees on the embedding's size and quality. The outlook is less clear when reducing the number of points $n$, and there are two separate paradigms that each provide distinct advantages.  On the one hand, we have uniform sampling, which runs in sublinear time but may miss important subsets of the data and therefore can only guarantee accuracy under certain strong assumptions on the data \cite{HJJ23}.  On the other hand, the most accurate sampling strategies provide the \emph{strong coreset} guarantee, wherein the cost of any solution on the compressed data is within an $\varepsilon$-factor of that solution's cost on the original dataset \cite{CSS21}.  

\paragraph{Our contributions}
We study both paradigms (uniform sampling and strong coresets) with respect to a classic problem -- compression for the $k$-means and $k$-median objectives. Whereas uniform sampling provides optimal speed but no worst-case accuracy guarantee, all available coreset constructions have a running time of at least $\tilde{\Omega}(nd+nk)$ when yielding tight bounds on the minimum number of samples required for accurate compression. 

It is easy to show that any algorithm that achieves a compression guarantee must read the entire dataset\footnote{A simple example is to consider a data matrix in $\mathbb{R}^{n \times d}$ with one single large entry and the remaining entries being very small. The algorithm must find this large entry to yield a good clustering.}. Thus a clear open question is what guarantees are achievable in linear or nearly-linear time. Indeed, currently available fast sampling algorithms for clustering \cite{bachem2018scalable, kmeans_sublinear_bachem16} cannot achieve strong coreset guarantees.  Recently, \cite{DSWY22} proposed a method for strong coresets that runs in time $\tilde{O}(nd + nk)$ and conjectured this to be optimal for $k$-median and $k$-means.

While this bound is effectively optimal for small values of $k$, there are many applications such as computer vision \cite{ExarchakisOL22} or algorithmic fairness \cite{Chierichetti0LV17} where the number of clusters can be larger than the number of features by several orders of magnitude. In such settings, the question of time-optimal coresets remains open. Since the issue of determining a coreset of optimal size has recently been closed \cite{CSS21,CLSSS22,huangLB}, this is arguably the main open problem in coreset research for center-based clustering.  We resolve this by showing that there exists an easy-to-implement algorithm that constructs coresets in $\tilde{O}(nd)$ time -- only logarithmic factors away from the time it takes to read in the dataset.

Nonetheless, this does not fully illuminate the landscape among sampling algorithms for clustering in practice. Although our algorithm achieves both an optimal runtime and an optimal compression, it is certainly possible that other, cruder methods may be just as viable for all practical purposes.  We state this formally in the following question: \textit{When are optimal $k$-means and $k$-median coresets necessary and what is the practical tradeoff between coreset speed and accuracy?}

To answer this, we perform a thorough comparison across the full span of sampling algorithms that are faster than our proposed method. Through this we verify
that, while these faster methods are sufficiently accurate on many real-world datasets, there exist data distributions that cause catastrophic failure for each
of them. Indeed, these cases can only be avoided with a strong-coreset method. Thus, while many practical settings do not require the full coreset guarantee,
one cannot cut corners if one wants to be confident in their compression. We verify that this extends to the streaming paradigm and applies to downstream
clustering approaches.

In summary, our contributions are as follows:
\begin{itemize}

    \item We show that one can obtain strong coresets for $k$-means and $k$-median in $\tilde{O}(nd)$ time. This resolves a conjecture on the necessary runtime
        for $k$-means coresets \cite{DSWY22} and is theoretically optimal up to log-factors.

    \item Through a comprehensive analysis across datasets, tasks, and streaming/non-streaming paradigms, we verify that there is a necessary tradeoff between
        speed and accuracy among the linear- and sublinear-time sampling methods. This gives the clustering practitioner a blueprint on when to use each
        compression algorithm for effective clustering results in the fastest possible time.

\end{itemize}

\section{Preliminaries and Related Work}
\label{sec:preliminaries}

\subsection{On Sampling Strategies.}
\label{ssec:sens_sampling}
As discussed, we focus our study on linear- and sublinear-time sampling strategies. Generally speaking, we consider compression algorithms through the lens of three requirements: 
\begin{itemize}
    \item Finding the compression should not take much longer than the time to read in the dataset.
    \item The size of the compressed data should not depend on the size of the original dataset.
    \item Candidate solutions on the compressed data are provably good on the original data.
\end{itemize}
If these requirements are satisfied then, when analyzing runtimes on large datasets, it is always preferable to compress the dataset and then perform the task in question on the compressed representation.

Specifically, given a dataset $P \in \R^{n \times d}$, we concern ourselves with sampling $\Omega \in \R^{m \times d} \subset P$ (such that $m \ll n$) along with a weight vector $w \in \R^m$. The goal is then that for any candidate solution $\calC$, $\Omega$ provides us with an idea of the solution's quality with respect to the original dataset, i.e. $\sum_{p \in \Omega} w_p \cost(p, \calC) \approx \sum_{p \in P} \cost(p, \calC)$ for a problem-specific cost function.

The quickest sampling strategy, running in sublinear time, is uniform sampling. It is clear, however, that this cannot provide any cost-preservation guarantee as missing a single extreme outlier will cause the sampling strategy to fail. Thus, any approach that outperforms uniform sampling must read in the entire dataset and therefore run in at least linear time\footnote{Indeed, sublinear algorithms always require some assumption on the input to provide guarantees, see \cite{Ben-David07,czumaj2007sublinear,HJJ23,meyerson2004k}.}. Among these more sophisticated sampling strategies, coresets offer the strongest compression guarantee:

\begin{definition}
    A \emph{strong $\eps$-coreset} is a pair $(\Omega, w)$ such that for \emph{any} candidate solution $\calC$, \[\sum_{p \in \Omega} w_p \cost(p, \calC) \in (1 \pm \eps) \cost(P, \calC).\] 
\end{definition} 

Going forward, we will discuss this in the context of the $k$-median and $k$-means cost functions: for dataset $P \in \R^d$ with weights $w : P \rightarrow \R^+$, and any $k$-tuple $\calC$ in $\R^d$, \[\cost_z(P, \calC) := \sum_{p \in P} w_p \dist^z(p, \calC),\] with $z=1$ for $k$-median and $z=2$ for $k$-means. We use $\opt$ to denote $\min_{\calC} \cost_z(P, \calC)$ and will denote an $\alpha$-approximation as any candidate solution
$\calC$ such that $\cost(\calC) \leq \alpha \cdot \opt$.

Recently, sampling with respect to sensitivity values has grown to prominence due to its simple-to-obtain coreset guarantee.  True sensitivity values are defined
as $\sup_{\calC} \frac{\dist^z(p, \calC)}{\cost_z(P, \calC)}$, where the supremum is taken over all possible solutions $\calC$. Intuitively, this is a measure
of the maximum impact a point can have on a solution and is difficult to evaluate directly.
Thus, the approximate sensitivity-sampling algorithm we consider is the following (as introduced in \cite{FL11}).
Given an $\alpha$-approximate solution $\calC$ to a clustering problem, importance scores are defined as
\begin{equation}
\label{eq:sensitivity}
\sigma_\calC(p) = \alpha \cdot \left( \dfrac{\cost(p, \calC)}{\cost(\calC_{p}, \calC)} + \dfrac{1}{|\calC_p|} \right),
\end{equation}
where $\calC_p$ is the cluster that $p$ belongs to. This is always an upper-bound on the sensitivity values \cite{FL11}. In essence, sampling enough points
proportionately to these values guarantees an accurate compression. The following paragraph discusses how the points must be re-weighted to guarantee the
coreset property:

The coreset $\Omega$ consists of $m$ points sampled proportionate to $\sigma$ with weights defined as follows: for any sampled point $p$, define
$w_p := \frac{1}{\Pr[p \in \Omega]} = \frac{\sum_{p'} \sigma_\calC(p')}{m \cdot \sigma_\calC(p)}$. The weights ensure that the cost estimator is unbiaised: in expectation, for any solution $C$, the cost evaluated on the sample should be equal to the original cost.
It was shown in \cite{HuangV20} that, when $\calC$ is a $O(1)$-approximation, sampling $m = \tilde O\lpar
k \eps^{-2z-2}\rpar$ many points was enough to ensure concentration, namely, $\Omega$ is a coreset with probability at least $2/3$ -- the conventional success probability for Monte-Carlo algorithms.

To perform this algorithm, the bottleneck in the running time lies in computing the solution $\calC$ as well as then obtaining costs of every point to its assigned center in $\calC$. This takes $\tilde O(nk + nd)$ time when using a bicriteria approximation algorithm\footnote{For $k$-means an $(\alpha,\beta)$ bicriteria approximation is an algorithm
that computes a solution $\calC$ satisfying $\cost(P, \calC)\leq \alpha\cdot \opt$ and $|\calC|\leq \beta\cdot k$.} such as the standard $k$-means++ algorithm
\cite{ArV07} combined with dimension reduction techniques (see for example \cite{BecchettiBC0S19,CohenEMMP15,MakarychevMR19}). This is precisely what was conjectured as the necessary runtime for obtaining $k$-means and $k$-median coresets, as merely assigning points to their
centers from the bicriteria seems to require $\Omega(nk)$ running time \cite{DSWY22}. 

\begin{figure}
    \centering
    \begin{tabular}{c}
        \includegraphics[width=.95\linewidth]{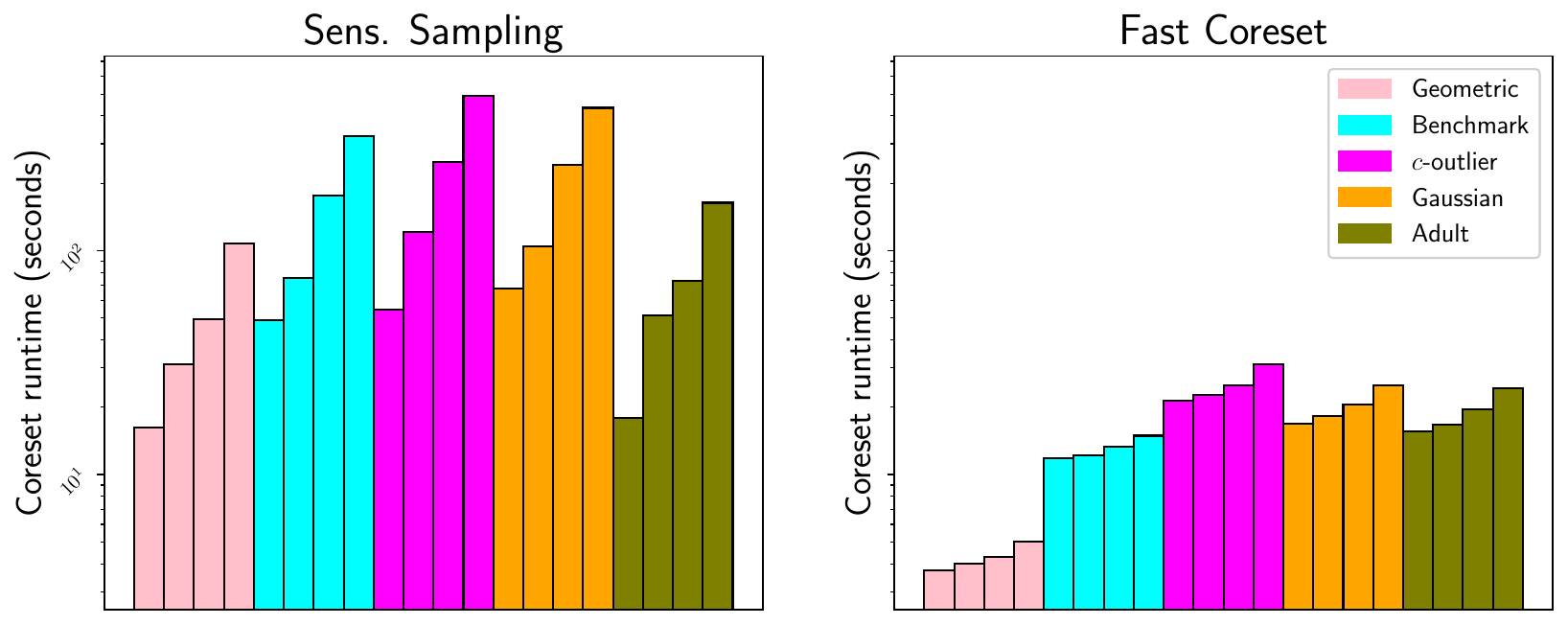}
    \end{tabular}
    \caption{Mean runtime over five runs as we vary $k$ for sensitivity sampling and Fast-Coresets. Bars are $k=50, 100, 200, 400$; y-axis is log-scale.}
    \label{fig:coreset_size_on_sens_quality}
\end{figure}

\subsection{Other Coreset Strategies}
\label{ssec:clustering_prelim}

Many of the advancements regarding coresets have sought the smallest coreset possible across metric spaces and downstream objectives. The most prominent
examples are in Euclidean space \cite{BadoiuHI02, HaM04, Chen09, HuangV20, stoc22}, with much of the focus on obtaining the optimal size in the $k$-means and
$k$-median setting. Recently, a lower bound \cite{huangLB} showed that the group sampling algorithm developed in \cite{CSS21, stoc22, CLSSS22} is optimal.

Although optimal coresets have size $\tilde{O}(k\cdot \varepsilon^{-2} \min(k^{z/(z+2)},\varepsilon^{-z}))$ \cite{CLSSS22} and are theoretically smaller than those obtained by sensitivity sampling, the experiments of \cite{chrisESA} showed that the latter is always more efficient in practice.

In terms of other linear-time methods with sensitivity sampling, we are only aware of the lightweight coresets approach \cite{bachem2018scalable}, wherein one
samples with respect to the candidate solution $\calC=\{\mu\}$, i.e. the mean of the data set, instead of a candidate solution with $k$ centers. This runs in $O(nd)$ time but provides a weaker guarantee -- one incurs an
additive error of $\varepsilon\cdot \cost(P,\{\mu\})$.  We note that this can be generalized to performing sensitivity sampling using a $\calC$ that has fewer
than $k$ centers. We discuss this in more depth in Section \ref{ssec:algorithms}.

Lastly, the BICO coreset algorithm \cite{bico} utilizes the SIGMOD test-of-time winning BIRCH \cite{birch} clustering method to obtain $k$-means coresets while the Streamkm++ method~\cite{streamkm++} uses $k$-means++~\cite{ArV07} to obtain a coreset whose size depends on $n$ and is exponential in $d$. While both were developed to perform quickly in a data-stream, we show in Section \ref{sec:results} that they do not provide strong coreset guarantees in practice for reasonable coreset sizes.

All efficient coreset constructions are probabilistic, making coresets difficult to evaluate. For example, it is co-NP-hard to check whether a candidate compression is a weak coreset \cite{chrisESA}\footnote{A weak coreset guarantee only requires that a $(1+\varepsilon)$ approximation computed on the coreset yields a $(1+\varepsilon)$ on the entire point set.}. Therefore, although coreset algorithms succeed with some high probability, it is unclear how to computationally verify this. We refer to \cite{chrisESA} for further discussion on this topic and discuss our evaluation metrics in Section \ref{sssec:metrics}.

\subsection{Coresets for Database Applications}
MapReduce\footnote{The MapReduce computation model is the following: there are several computation entities, each with memory constraints that prevent them from holding the entire dataset. The costly operation is in exchanging data between those entities, implying that the complexity is measured in terms of rounds of communication and size of the memory exchanged.} is one of the most popular architectures for large scale data analysis (see for example \cite{ChierichettiDK14, CordeiroTTLKF11,EneIM11} and references therein for MapReduce algorithms for clustering).
Within this context, strong coresets are `embarrassingly parallel' and have a natural synergy with the MapReduce framework due to the following two properties. First, if two coresets are computed on subsets of data then their union is a coreset for the union of the data subsets. Second, many coreset constructions produce a compression with size completely independent of $n$, allowing the coreset to be stored in memory-constrained environments.

Using these two properties, one can compress the data in a single round of MapReduce as follows~\cite{coreset_survey_mapreduce}. The data is partitioned randomly among the $m$ entities, each of which computes a coreset  and sends it to a central host. By the first property, the union of these coresets is a coreset for the full dataset. Thus, the host now possesses a small summary of the full dataset with strong coreset guarantees. By the second property, this summary's size does not depend on the size of the original dataset -- in particular, the total size of the messages received by the host is independent of $n$. Lastly, by the coreset guarantee, any solution that is good w.r.t. the summary is good w.r.t. the original data (and vice versa).

This idea allows one to compute an approximate solution to $k$-means and $k$-median, and do so efficiently, in MapReduce: each computation entity needs to communicate merely $O(k)$ bits, where $k$ is the number of clusters. The computational burden can therefore be completely parallelized up to the time required to compute a coreset in each entity -- precisely the focus of this paper. We explore similar aggregation strategies experimentally in Section~\ref{ssec:streaming}.

\subsection{Quadtree Embeddings}

A common techniques for designing fast algorithms is to embed the Euclidean space into a tree metric using \emph{quadtree embeddings}.  The central idea is that any hypercube in the input space can be split into $2^d$ sub-cubes. We can represent this in a tree-structure, where each sub-cube has the original hypercube as its parent. Centering randomly the original hypercube, and appropriately setting the weight of each branch then preserves the expected distance between points in different sub-cubes within an $O(\sqrt{d} \log \Delta)$ factor. Here, $\Delta$ is the \emph{spread} of the point set and is equal to the ratio of the largest distance over the smallest non-zero distance.  Given this context, we now introduce quadtree embeddings more formally:

\paragraph*{Formal Overview}

Given a set of points in $\mathbb{R}^{n \times d}$, we want to return a tree-structure that roughly preserves their pairwise distances.  To do this, our first
step is to obtain a box enclosing all input points, centered at zero, with all side lengths equal to $\Delta$. This can be done as follows: select an arbitrary
input point, and translate the dataset so that this point is at the origin. Then, using $O(nd)$ time, set $\Delta$ to be the maximum distance from any point to
the origin. Note that, up to rescaling the points so that the smallest distance equals $1$, this is equivalent to the \emph{spread} as described in the previous
paragraph.

Having obtained this box, add a shift $s$ (picked uniformly at random in $[0, \Delta]$) to all the points' coordinates so that the input is now in the box
$[-2\Delta, 2\Delta]^d$. This transformation does not change any distances and therefore preserves the $k$-median and $k$-means costs.  The $i$-th level of the
tree (for $i \in \lbra 0, ..., \log \Delta \rbra$) is constructed by centering a grid of side length $2^{-i} \cdot 2\Delta$ at $0$, making each grid-cell a node
in the tree.  The parent of a cell $c$ is simply the cell at level $i-1$ that contains $c$, and the distance between $c$ and its parent is set to $2^{-i} \cdot
2\Delta \cdot \sqrt{d}$ (the length of the hypercube's diagonal). This embedding takes $O(nd \log \Delta)$ time to construct, where $\log \Delta$ is the depth
of the tree\footnote{Crucially, note that our \cref{alg:crudeApx} does not need to build the full tree. Instead, we only require $\log \log \Delta$
levels of it, resulting in our improved complexity.}. The linearity in the $\log \Delta$ term comes from the fact that this is the maximum depth of the tree.

The distortion of this embedding is at most $O(d \log \Delta)$, as stated in the following lemma:
\begin{lemma}[Lemma 11.9 in \cite{har2011geometric}]\label{lem:quadtreeDist}
The distances in the tree metric $d_T$ satisfy
$\forall p,q, \|p-q\| \leq \mathbb{E}[d_T(p, q)] \leq O(d \log \Delta) \|p-q\|$, where the expectation is taken over the random shift $s$ of the decomposition.
\end{lemma}

A simple proof (and further intuition on quadtree embeddings) can be found in \cite{har2011geometric}. The result follows from combining linearity
of expectation and the fact that two points $p$ and $q$ are separated at level $i$ with probability at most $\sqrt{d} \|p-q\| \frac{2^i}{\Delta}$ (as in the
proof of \cref{lem:quadtreeSep}).

\section{Fast-Coresets}
\label{sec:theory_0}

\begin{algorithm}[tb]
   \caption{Fast-Coreset($P, k, \eps, m$)}
   \label{alg:main}
\begin{algorithmic}[1]
   \State {\bfseries Input:} data $P$, number of clusters $k$, precision $\eps$ and target size $m$
   \State Use a Johnson-Lindenstrauss embedding to embed $\tilde P$ of $P$ into $\tilde d = O(\log k)$ dimensions
   \State Find approx. solution $\tilde \calC = \lbra \tilde c_1, ..., \tilde c_k\rbra $ on $\tilde P$ and assignment $\tilde \sigma : \tilde P \rightarrow
   \tilde \calC$ by \fkmeans.	
   \State Let $\calC_i = \tilde \sigma^{-1}(c_i)$. Compute the $1$-median (or $1$-mean) $c_i$ of each $\calC_i$ in $\R^d$.
   \State For each point $p \in \calC_i$, define
   $s(p) = \frac{\dist^z(p, c_i)}{\cost(\calC_i, c_i)}+ \frac{1}{|\calC_i|}$.
   \State Compute a set $\coreset$ of $m$ points randomly sampled from $P$ proportionate to $s$.
   \State For each $\calC_i$, define $|\hat \calC_i|$ the estimated weight of $\calC_i$ by $\coreset$, namely $|\hat \calC_i| := \sum_{p \in \calC_i \cap
   \coreset} \frac{\sum_{p' \in P}s(p')}{s(p)m}$.
   \State {\bfseries Output:} the coreset $\coreset$, with weights $w(p) = \frac{\sum_{p' \in P}s(p')}{s(p)m} \lpar (1+\eps)|\calC_i| - |\hat \calC_i|\rpar$
\end{algorithmic}
\end{algorithm}

\paragraph{Technical Preview.}

We start by giving an overview of the arguments in this section.

There exists a strong relationship between computing coresets and approximate solutions -- one can quickly find a coreset given a reasonable solution and vice-versa. Thus, the general blueprint is as follows: we very quickly find a rough solution which, in turn, facilitates finding a coreset that approximates all solutions. Importantly, the coreset size depends linearly on the quality of the rough solution. Put simply, the coreset can compensate for a bad initial solution by oversampling. Our primary focus is therefore on finding a sufficiently good coarse solution quickly since, once this has been done, sampling the coreset requires linear-time in $n$. Our theoretical contribution shows how one can find this solution on Euclidean data using dimension reduction and quadtree embeddings.

\paragraph{Formal Results.} In this section, we first combine two existing results to produce a strong coreset in time $\tilde{O}(nd \log \Delta)$, where $\Delta$ is the spread of
the input.  We show afterwards how to reduce the dependency in $\Delta$ to $\log \log \Delta$, giving the desired nearly-linear runtime.

Our method is based on the following observations about the group sampling \cite{CSS21} and sensitivity sampling \cite{FL11} coreset construction algorithms. Both start by computing a solution $\calC$. When $\calC$ is a $c$-approximation, they compute a coreset with distortion $1 \pm c \eps$ of size $\tilde O\lpar k \eps^{-z-2}\rpar$ and $\tilde O\lpar k \eps^{-2z -2}\rpar$, respectively. Hence, by rescaling $\eps$, they provide an $\eps$-coreset with size $\tilde O\lpar k (\eps/c)^{-z-2}\rpar$ and $\tilde O\lpar k (\eps/c)^{-2z-2}\rpar$.  This leads to the following fact:

\begin{fact}
    \label{fact:logApprox}
    [See Theorem 1 in \cite{CSS21} and 4.8 in \cite{FL11}\footnote{The references use an $O(1)$-approximation, but the proofs work with an $O\lpar \log^{O(1)} k\rpar$ approximation, with an added $O\lpar \log^{O(1)} k\rpar$ factor in the coreset size.}] Let $\calC$ be an $O\lpar \log^{O(1)} k\rpar$ approximation to $k$-median or $k$-means.  Then group (resp. sensitivity) sampling using solution $\calC$ computes a coreset of size $\tilde O\lpar k \eps^{-z-2}\rpar$ (resp. $\tilde O\lpar k \eps^{-2z-2}\rpar$) in time $\tilde O(nd)$.
\end{fact}

To turn \cref{fact:logApprox} into an algorithm, we use the quadtree-based \fkmeans approximation algorithm from \cite{cohen2020fast}, which has two key properties: 
\begin{enumerate}
    \item \fkmeans runs in $\tilde O\lpar n d \log \Delta\rpar$ time (Corollary 4.3 in \cite{cohen2020fast}), and
    \item \fkmeans computes an assignment from input points to centers that is an $O\lpar d^z \log k\rpar$ approximation to $k$-median ($z=1$) and $k$-means ($z=2$) (see Lemma 3.1 in \cite{cohen2020fast} for $z=2$ and the discussion above that lemma for $z=1$). Applying dimension reduction techniques \cite{MakarychevMR19}, the dimension $d$ may be replaced by a $\log k$ in time $\tilde O(nd)$. This results in a $O\lpar\log^{z+1} k\rpar$ approximation.
\end{enumerate}

The second property is crucial for us: the algorithm does not only compute centers, but also assignments in $\tilde{O}(nd\log \Delta)$ time. Thus, it satisfies the requirements of Fact \ref{fact:logApprox} as a sufficiently good initial solution. We describe how to combine \fkmeans with sensitivity sampling in \cref{alg:main} and prove in \cref{ssec:cor_proof} that this computes an $\eps$-coreset in time $\tilde O(nd \log \Delta)$:

\begin{corollary}\label{cor:mainAlg}
\cref{alg:main} runs in time $\widetilde O\lpar n d \log \Delta\rpar$ and computes an $\eps$-coreset for $k$-means.
\end{corollary}
Furthermore, we generalize \cref{alg:main} to other fast $k$-median approaches in \cref{app:extensions}.

Thus, one can combine existing results to obtain an $\eps$-coreset without an $\widetilde{O}(nk)$ time-dependency.  However, this has only replaced the $\widetilde{O}(nd + nk)$ runtime by $\widetilde{O}(nd \log \Delta)$. Indeed, the spirit of the issue remains -- this is not near-linear in the input size.

We verify that $\log \Delta$ is on the same order as $n$ by devising a dataset that has $n - n'$ points uniformly in the $[-1, 1]^2$ square. Then, for $r \in \mathbb{Z}^+$, we produce a sequence of
points at $(0, 1), (0, 0.5), \cdots, (0, 0.5^r)$ and copy this sequence $n' / r$ times, each time with a different $x$ coordinate. The result is a dataset of
size $n$ where $\log \Delta$ grows linearly with $r \in o(n)$. The resulting linear time-dependency can be seen in Table \ref{tbl:logdelta}.

\begin{table}
    \centering
    \begin{tabular}{ccccc}
        \hline \\
        \vspace*{-0.65cm} \\
        \vspace*{0.05cm}
        Value of $r$ & 20 & 30 & 40 & 50 \\
        \vspace*{0.05cm}
        Runtime & 13.5 $\pm$ 0.16 & 14.2 $\pm$ 0.33 & 15.5 $\pm$ 0.02 & 16.2 $\pm$ 0.16 \\
        \hline
    \end{tabular}
    \captionof{table}{Mean runtime in seconds for \fkmeans as a function of $r \sim \log \Delta$, taken over five runs.}
    \label{tbl:logdelta}
\end{table}

\section{Reducing the Impact of the Spread}
\label{sec:theory} 
\newcommand{\boxsize}{\textsc{MaxDist}}
We now show how one can replace the linear time-dependency on $\log \Delta$ with a logarithmic one (i.e., $\log \Delta \rightarrow \log (\log \Delta)$).

\paragraph*{Overview of the Approach}

Without loss of generality, we assume that the smallest pairwise distance is at least $1$, and $\Delta$ is a (known) upper-bound on the diameter of the input. To remove
the $\log\Delta$ dependency, we proceed by producing a substitute dataset $P'$ that has a lower-bound on its minimum distance and an upper-bound on its maximum
distance. We then show that, with overwhelming likelihood, reasonable solutions on $P'$ have the same cost as solutions on $P$ up to an additive error.

In order to produce the substitute dataset $P'$, we first find a crude upper-bound on the cost of the optimal solution to the clustering task on $P$. We
then create a grid such that, for every cluster in the optimal solution, the cluster is entirely contained in a grid cell with overwhelming likelihood -- in some sense, points in different grid cell do not interact. We can then freely move these
grid cells around without affecting the cost of the solution, as long as they stay far enough away so that points in different cells still do not interact. 
Thus, we can constrain the diameter of $P'$ to be small with respect to the quality of the approximate solution. We show later how to constrain the minimum distance to be comparable to the quality of this approximate solution as well.


We will focus this section on the simpler $k$-median problem but show how to reduce $k$-means to this case in \cref{app:redKM}.

\subsection{Computing a crude upper-bound}
\label{ssec:crude_bound}

As described, we start by computing an approximate solution $U$ such that $\opt \leq U \leq \poly(n) \cdot \opt$. For this, the first step is to embed the input
into a quadtree. This embedding has two key properties. First, distances are preserved up to a multiplicative factor $O(d \log \Delta)$, and therefore the
$k$-median cost is preserved up to this factor as well. Second, the metric is a \emph{hierarchically separated tree}: it can be represented with a tree, where
points of $P$ are the leafs. The distance between two points is then given by the depth of their lowest common ancestor -- if it is at depth $\ell$, their distance
is $\sqrt{d} \Delta 2^{-\ell}$.  Our first lemma shows that finding the first level of the tree for which the input lies in $k+1$ disjoint subtrees provides us
with the desired approximation. 

\begin{lemma}\label{lem:apxTree} [Proof in \cref{app:apx-tree-proof}]
Let $\ell$ be the first level of the tree with at least $k+1$ non-empty subtrees. Then, $\sqrt{d}2^{-\ell+1} \cdot \Delta \leq
\opt_T \leq n \cdot \sqrt{d}2^{-\ell+4} \cdot \Delta$, where $\opt_T$ is the optimal $k$-median solution in the tree metric.

\end{lemma}

We prove this in \cref{app:apx-tree-proof}. A direct consequence  is that the first level of the tree for which at least $k+1$ cells are non empty allows us to
compute an $O(n)$-approximation for $k$-median on the tree metric. Since the tree metric approximates the oringial Euclidean metric up to $O(d \log
\Delta)$, this is therefore an $O(n d \log \Delta)$-approximation to $k$-median in the Euclidean space.

To turn this observation into an algorithm, one needs to count the number of non-empty cells at a given level $\ell$: for each point, we identify the cell that
contains it using modulo operations. Furthermore, we count the number of distinct non-empty cells using a standard dictionary data structure. This is done in
time $\tilde O(nd)$, and pseudo-code is given \cref{alg:crudeApx}. Using a binary search on the $O(\log \Delta)$ many levels then gives the following result:

\begin{lemma}\label{lem:crudeApx}[Proof in \cref{app:apx-tree-proof}]
There is an algorithm running in time $\tilde O(nd \log \log \Delta)$ that computes an $O(n d \log \Delta)$-approximation to $k$-median, and $O(n^3 d^2 \log^2
\Delta)$-approximation to $k$-means.
\end{lemma}

Given this crude approximate solution, it remains to create a substitute dataset $P'$ that
satisfies two requirements:
\begin{enumerate}
    \item First, $P'$ must have spread linear in the quality of the approximate solution. If this holds, Algorithm~\ref{alg:main} on $P'$ will take
        $\tilde{O}(nd \log \log \Delta)$ time.
    \item Second, any reasonable solution on $P'$ should be roughly equivalent to a corresponding solution on $P$. This would mean that running
        Algorithm~\ref{alg:main} on $P'$ gives us a valid coreset for $P$.
\end{enumerate}

\subsection{From Approximate Solution to Reduced Spread}
\label{ssec:reduce_spread}

Let $U$ be an upper-bound on the optimal cost, computed via \cref{lem:crudeApx}. We place a grid with side length $r:= \sqrt{d} n^2\cdot U$, centered at
a random point in $\{0, ..., r\}^d$.  The following lemma ensures that with high probability (over the randomness of the center of the grid), no cluster of the
optimal solution is spread on several grid cells.

\begin{lemma}\label{lem:quadtreeSep}

The probability over the center's location that two points $p$ and $q$ are in different grid cells is at most $\frac{\|p-q\|}{n^2 U}$

\end{lemma}
\begin{proof}
We first bound the probability that there is a grid line along the $i$-th dimension between $p$ and $q$. Let $p_i, q_i$ be the $i$-th coordinate of $p$ and $q$,
assume $p_i \leq q_i$ and let $\ell \in \mathbb{Z}$ such that $p_i - \ell r \in [0, r)$.  Then, $p$ and $q$ are separated along dimension $i$ if and only if the
$i$-th coordinate of the center is in $[p_i - \ell r, q_i - \ell r]$. This happens with probability $|p_i - q_i|/r$.  Finally, a union-bound over all
coordinates shows that $p$ and $q$ are in different grid cells with probability at most $\sum |p_i-q_i| / r \leq \sqrt{d} \|p-q\|/r$.
\end{proof}

Since $U$ is larger than the distance between any input point and its center in the optimal solution, a union-bound ensures that with probability $1-1/n$, no
cluster of this solution is split among different cells.  In particular, there are at most $k$-non empty cells which we can identify using a dictionary.  We
call these ``boxes".
Each box $j$ has a middle point, which we call $m_j$.

From this input, we build a new set of points $P'$ as follows.  For each dimension $i \in \lbra 1, ..., d \rbra$, sort the $k$ boxes according to their value on
dimension $i$. Then, for each $j \in \lbra 1, ..., k\rbra$, let $m^i_j$  be the $i$-th coordinate of the middle-point of the $j$-th box. If $m^i_{j+1} - m^i_j
\geq 2r$, then for all boxes $j'$ with $j' > j$, shift the points of $j'$ by $m^i_{j+1} - m^i_j - 2r$ in the $i$-th dimension. This can be implemented in
near-linear time, as described in \cref{alg:reduce-diam} (presented in \cref{app:pseudoCode}). In essence, we take boxes that are far apart and bring them
closer together. The dataset $P'$ obtained after these transformations has the following properties:

\begin{proposition}\label{prop:boxes}
    Let $P'$ be the dataset produced by Algorithm~\ref{alg:reduce-diam}. It holds that:
    \begin{enumerate}
        \item in $P'$, the diameter of the input is $O(d n^2\cdot U \cdot k)$ with probability $1-1/n$ over the randomness of the grid, and
        \item two boxes that are adjacent (respectively non-adjacent) in $P$ are still adjacent (resp. non-adjacent) in $P'$.
    \end{enumerate}
\end{proposition}
\begin{proof}

By construction, the max distance between the middle-points of any two boxes in $P'$ is $2r = 2\sqrt d n^2\cdot U$. \cref{lem:quadtreeSep} ensures that, with probability $1-1/n^2$, any two points in the same cluster of the optimal solution (e.g., at distance less than $\opt \leq U$ from each other) are in the same box. Therefore, a union-bound ensures that there are at most $k$ boxes.  It then follows that the total distance along a coordinate is at most $2kr$, and the diameter of the whole point set is $\sqrt{d} \cdot 2kr$.

If two boxes were adjacent in $P$, then along any dimension their middle-points have distance at most $r$. Thus, if one of the adjacent boxes moves along any
dimension, the other must as well and they will remain adjacent.  If they are not adjacent, there is at least one dimension where their middle-points are at
distance at least $2r$ from one another. Along each such dimension, their shift cannot bring them closer than to within $2r$ of one another and they will stay
non-adjacent.

\end{proof}

The first property allows us to reduce the spread to $(nd \log \Delta)^{O(1)}$.  Indeed, one can round the coordinates of every point in $P'$ to the closest
multiple of $g := \frac{U}{n^4 d^{2} \log \Delta}$.  Combined with the diameter reduction, this ensures that the spread of the dataset obtained is at most $(nd
\log \Delta)^{O(1)}$.  Furthermore, the second property of \cref{prop:boxes} combined with the choice of $g$ ensures that the cost of any reasonable solution is
the same before and after the transformation, as stated in the following lemma:

\begin{lemma}\label{lem:reduceSpread}

Let $P'$ be the outcome of the diameter reduction and rounding steps. With probability $1-1/n$ over the randomness of the grid, $P'$ has spread $(nd \log
\Delta)^{O(1)}$.

Suppose $U$ is such that $\opt \leq U \leq d n^2 \opt$, and let $\calS'$ be a solution for $k$-median (resp. $k$-means) on $P'$ with cost at most $d n^2
\opt$ (resp. $d^2 n^4 \opt$ for $k$-means). Then, one can compute a $k$-median solution for $P$, with the same cost as $\calS'$ for $P'$ up to an
additive error $\opt / n$, in time $O(nd)$. This also works if we replace $P'$ and $P$.

\end{lemma}
\begin{proof}
First, in rounding points to the closest multiple of $g$, the distance between any point and its rounding is at most $g \leq \frac{U}{n^4 d^2 \log \Delta} \leq
\frac{\opt}{n^2}$. Summing over all points, any solution computed on the gridded data has cost within an additive factor $\pm \frac{\opt}{n}$ of the true cost. 


Let $\calS$ be the solution obtained from $\calS'$ by reversing the construction of $P'$, namely re-adding the shift that was substracted to every box. Since
adjacency is preserved by Proposition~\ref{prop:boxes}, all points that are in the same cluster have the same shift, and therefore all intra-cluster distances
are the same in $P$ and $P'$.  Therefore, the costs are equal and, by extension, $\cost \left( \calS \right) \in \cost \left( \calS' \right) \pm \opt / n$,
where the additive $\opt /n$ comes from the rounding.

Finally, the smallest non-zero distance is $g = \frac{U}{n^4 d^{2} \log \Delta}$, and with probability $1-1/n$ the diameter is $\sqrt{d} \cdot 3d n^2\cdot
U \cdot k$ (see \cref{prop:boxes}), implying that the spread of $P'$ is $(nd \log \Delta)^{O(1)}$.
\end{proof}

Combining the algorithm of \cref{lem:crudeApx}, which gives a bound on $U$, with \cref{lem:reduceSpread} brings us to the following theorem:

\begin{theorem}
\label{thm:main_theorem}

Given $P \subset \R^d$ with spread $\Delta$, there is an algorithm running in time $O(nd \log \log \Delta)$ that computes a set $P'$ such that (1) with
probability $1-1/n$, the spread of $P'$ is $\poly(n,d, \log(\Delta))$ and (2) any solution with cost at most $d n^2 \opt$ for  $k$-median (resp. $d^2 n^4 \opt$
for $k$-means) on $P'$ can be converted in time $O(nd)$ into a solution with same cost on $P$, up to an additive error $O(\opt / n)$.

\end{theorem}

To summarize, we have shown that one can, in $\tilde{O}(nd)$ time, find a modified dataset $P'$ that preserves the cost of corresponding $k$-means and
$k$-median solutions from $P$. Importantly, this $P'$ has spread that is logarithmic in the spread of $P$. As a result, one can apply Algorithm \ref{alg:main}
onto $P'$ in $\tilde{O}(nd)$ time without compromising the compression quality with respect to $P$. Lastly, this compression on $P'$ can be re-formatted onto
$P$ in $\tilde{O}(nd)$ time.

\section{Fast Compression in Practice}
\label{sec:results}

Despite the near-linear time algorithm described in Sections~\ref{sec:theory_0} and~\ref{sec:theory}, the coreset construction of \cref{alg:main} nonetheless requires a bounded approximation to the clustering task before the sampling can occur. Although theoretically justified, it is unclear how necessary this is in practice -- would a method that cuts corners still obtain good practical compression?

\paragraph*{Metrics}
\label{sssec:metrics}

To answer this question, we analyze the sampling methods along two metrics -- compression accuracy and construction time. Although measuring runtime is standard, it is unclear how to confirm that a subset of points satisfies the coreset property over all solutions. To this end, we use the distortion measure introduced in~\cite{chrisESA} $\max \left( \dfrac{\cost(P, \calC_{\Omega})}{\cost(\Omega, \calC_{\Omega})}, \dfrac{\cost(\Omega, \calC_{\Omega})}{\cost(P, \calC_{\Omega})} \right),$ where $\calC_{\Omega}$ is a candidate solution computed over the coreset $\Omega$. This will be within $1+\varepsilon$ if the coreset guarantee is satisfied but may be unbounded otherwise.  We refer to this as the \emph{coreset distortion}.

\subsection{Goal and Scope of the Empirical Analysis}
\label{ssec:intro_experiments}
To motivate the upcoming experiments, we begin by asking ``how do other sampling strategies compare to standard sensitivity sampling?'' For this preliminary experiment, we focus on the uniform sampling and Fast-Coreset algorithm (Algorithm~\ref{alg:main}). For each, we evaluate its distortion across the following real datasets: Adult~\cite{Dua:2019}, MNIST~\cite{mnist}, Taxi~\cite{taxi}, Star~\cite{star}, Song~\cite{song}, Census~\cite{census}, and Cover Type~\cite{covtype}. The dataset characteristics are summarized in Table~\ref{tbl:datasets}.

The resulting comparisons can be found in Table~\ref{tbl:motivating_table}. Since sensitivity sampling is the recommended coreset method~\cite{chrisESA}, we use it to obtain a baseline distortion for each dataset\footnote{We use $k=100$ and coreset size $m=40k$ for these experiments.}. We then compare uniform sampling and Fast-Coresets against this baseline by showing the ratio of their distortion divided by the distortion obtained by sensitivity sampling. As guaranteed by Section~\ref{sec:theory}, we see that Fast-Coresets obtain consistent distortions with standard sensitivity sampling. However, the question is more subtle for uniform sampling -- it performs well on most real-world datasets but catastrophically fails on a few (for example, it is $\sim600\times$ worse on the taxi dataset when compared with standard sensitivity sampling).

This confirms that uniform sampling is not unequivocally reliable as an aggregation method -- although it is fast, it has the potential to miss important data points. On the other end of the runtime vs. quality spectrum, Fast-Coresets consistently provide an accurate compression but, despite being the fastest coreset method, are still significantly slower than uniform sampling. Thus, the fundamental question is: when is one safe to use fast, inaccurate sampling methods and when is the full coreset guarantee necessary?

We focus the remainder of the experimental section on this question. Specifically, we define a suite of compression methods that interpolate between uniform sampling and Fast-Coresets and evaluate these methods across a set of synthetic datasets. This allows us to experimentally verify the full spectrum of speed vs. accuracy tradeoffs and provide insight into which dataset characteristics are necessary before one can apply suboptimal sampling methods. 

We complement this with a comprehensive study of related topics, such as additional analysis on real datasets, comparing against other state-of-the-art coreset methods, and verifying that these results extend to the streaming setting. Unless stated otherwise, our experimental results focus on the $k$-means task.

\begin{minipage}{0.55\textwidth}
    \begin{tabular}{lcc}
        Dataset & \makecell{$\dfrac{\text{Uniform Dist.}}{\text{Sensitivity Dist.}}$} & \makecell{$\dfrac{\text{Fast-Coreset Dist.}}{\text{Sensitivity Dist.}}$} \\
        \hline
        Adult & 1.00 & 1.07 \\ 
        MNIST & 1.03 & 1.02 \\ 
        Star & \textbf{8.46} & 1.09 \\ 
        Song & 1.11 & 1.47 \\ 
        Cover-Type & 1.02 & 1.27 \\ 
        Taxi & \textbf{614} & 1.90 \\ 
        Census & 1.09 & 1.08 \\
        \hline
    \end{tabular}
    \captionof{table}{Ratio of each algorithms' distortion with respect to sensitivity sampling's distortion. Failure cases are bolded.}
    \label{tbl:motivating_table}
\end{minipage}
\quad \quad
\begin{minipage}{0.32\textwidth}
\vspace*{0.38cm}
    \begin{tabular}{lrr}
        Dataset & Points & Dim \\
        \hline
        \emph{Adult} & 48\,842 & 14 \\
        \emph{MNIST} & 60\,000 & 784 \\
        \emph{Star} & 138\,500 & 3 \\
        \emph{Song} & 515\,345 & 90 \\
        \emph{Cover Type} & 581\,012 & 54 \\
        \emph{Taxi} & 754\,539 & 2 \\
        \emph{Census} & 2\,458\,285 & 68 \\
        \hline
    \end{tabular}
    \captionof{table}{Description of real world datasets}
    \label{tbl:datasets}
\end{minipage}

\subsection{Experimental Setup}

\paragraph*{Algorithms}
\label{ssec:algorithms}

We compare Fast-Coresets (Algorithm~\ref{alg:main}) against 4 different benchmark sampling strategies that span the space between optimal time and optimal accuracy, as well as state-of-the-art competitors BICO~\cite{bico} and Streamkm++\cite{streamkm++}.
\begin{description}
        \item \emph{- Standard uniform sampling}. Each point is sampled with equal probability and weights are set to $n / m$, where $m$ is the size of the sample.
        \item \emph{- Lightweight coresets \cite{bachem2018scalable}}. Obtain a coreset by sampling sensitivity values with respect to the $1$-means solution,
            i.e. sensitivities are given by $\hat{s}(p) = 1/|P| + \cost(p, \mu) / \cost(P, \mu)$, where $\mu$ is the dataset mean.
        \item \emph{- Welterweight coresets}. For any $j \in \{1,..., k\}$, we compute a coreset using sensitivity sampling with respect to a candidate
            $j$-means solution.
        \item \emph{- Standard sensitivity sampling \cite{LS10}}. We refer to Section~\ref{sec:preliminaries}.
        \item \emph{- BICO \cite{bico}}. Utilizes BIRCH \cite{birch} to produce a $k$-means coreset in a stream.
        \item \emph{- Streamkm++ \cite{streamkm++}}. Uses $k$-means++ to create large coresets for the streaming $k$-means task.
\end{description}
We do not compare against group sampling~\cite{CSS21} as it uses sensitivity sampling as a subroutine and is merely a preprocessing algorithm designed to facilitate the theoretical analysis. By the authors' own admission, it should not be implemented.

\begin{table*}
    \centering
    \fontsize{7pt}{7pt}\selectfont
    \renewcommand{\arraystretch}{1.15}
    
    \hspace*{-1.25cm}
    \begin{tabular}{|c|cc|cc|cc|cc|}
        \hline
        & \multicolumn{8}{c|}{Method} \\
        \cline{2-9} & \multicolumn{2}{c|}{Uniform Sampling} & \multicolumn{2}{c|}{Lightweight} & \multicolumn{2}{c|}{Welterweight} & \multicolumn{2}{c|}{Fast Coreset} \\
        & $m=40k$ & $m=80k$ & $m=40k$ & $m=80k$ & $m=40k$ & $m=80k$ & $m=40k$ & $m=80k$ \\
        \cline{2-9}
        $c$-outlier & \underline{\textbf{405 $\pm$ 24K}} & \underline{\textbf{159 $\pm$ 20K}} & 1.07 $\pm$ 0.0 & \textbf{5.63 $\pm$ 84.6} & \textbf{9.44 $\pm$ 105} & 1.03 $\pm$ 0.0 & 1.12 $\pm$ 0.0 & 1.05 $\pm$ 0.0 \\
        Geometric & \underline{\textbf{86.3 $\pm$ 8.4K}} & \underline{\textbf{21.8 $\pm$ 652}} & 1.05 $\pm$ 0.0 & \underline{\textbf{11.6 $\pm$ 452}} & 1.11 $\pm$ 0.0 & 1.03 $\pm$ 0.0 & 1.11 $\pm$ 0.0 & 1.05 $\pm$ 0.0 \\
        Gaussian Mix. & 3.17 $\pm$ 3.42 & 1.43 $\pm$ 0.25 & 2.64 $\pm$ 1.61 & 1.81 $\pm$ 0.58 & 2.26 $\pm$ 1.55 & 1.28 $\pm$ 0.07 & 1.24 $\pm$ 0.0 & 1.13 $\pm$ 0.0 \\
        \makecell{Benchmark\\---} & \makecell{1.07 $\pm$ 0.0\\---} & \makecell{1.03 $\pm$ 0.0\\---} & \makecell{1.11 $\pm$ 0.0\\---} & \makecell{1.05 $\pm$
        0.0\\---} & \makecell{1.10 $\pm$ 0.0\\---} & \makecell{1.04 $\pm$ 0.0\\---} & \makecell{1.15 $\pm$ 0.0\\---} & \makecell{1.06 $\pm$ 0.0\\---} \\

        MNIST & 1.08 $\pm$ 0.0 & 1.03 $\pm$ 0.0 & 1.08 $\pm$ 0.0 & 1.03 $\pm$ 0.0 & 1.08 $\pm$ 0.0 & 1.04 $\pm$ 0.0 & 1.08 $\pm$ 0.0 & 1.04 $\pm$ 0.0 \\
        Adult & 1.08 $\pm$ 0.0 & 1.04 $\pm$ 0.0 & 1.09 $\pm$ 0.0 & 1.04 $\pm$ 0.0 & 1.32 $\pm$ 0.0 & 1.17 $\pm$ 0.0 & 1.17 $\pm$ 0.0 & 1.07 $\pm$ 0.0 \\
        Star & \textbf{5.43 $\pm$ 0.99} & 2.25 $\pm$ 0.12 & 1.94 $\pm$ 0.1 & 1.68 $\pm$ 0.0 & 1.85 $\pm$ 0.03 & 1.86 $\pm$ 0.03 & 1.13 $\pm$ 0.01 & 1.14 $\pm$ 0.01 \\
        Song & 1.30 $\pm$ 0.0 & 1.14 $\pm$ 0.0 & 1.13 $\pm$ 0.0 & 1.12 $\pm$ 0.0 & 1.18 $\pm$ 0.0 & 1.16 $\pm$ 0.0 & 1.50 $\pm$ 0.0 & 1.29 $\pm$ 0.0 \\
        Cover-Type & 1.12 $\pm$ 0.0 & 1.05 $\pm$ 0.0 & 1.11 $\pm$ 0.0 & 1.06 $\pm$ 0.0 & 1.17 $\pm$ 0.0 & 1.08 $\pm$ 0.0 & 1.11 $\pm$ 0.0 & 1.04 $\pm$ 0.0 \\
        Taxi & \underline{\textbf{3.7K $\pm$ 43K}} & \underline{\textbf{2.3K $\pm$ 12K}} & 2.4 $\pm$ 0.1 & 2.51 $\pm$ 0.05 & \textbf{5.82 $\pm$ 10.8} & \underline{\textbf{84.3 $\pm$ 263}} & 4.64 $\pm$ 0.09 & 4.90 $\pm$ 0.35 \\
        Census & 1.15 $\pm$ 0.0 & 1.07 $\pm$ 0.0 & 1.12 $\pm$ 0.0 & 1.06 $\pm$ 0.0 & 1.14 $\pm$ 0.0 & 1.06 $\pm$ 0.0 & 1.13 $\pm$ 0.0 & 1.07 $\pm$ 0.0 \\
        \hline
    \end{tabular}
    \caption{Distortion means and variances for different sample sizes across datasets for $k$-means; taken over 5 runs. Failure cases (distortion $>5$) are bolded. Catastrophic failures (distortion $>10$) are underlined.}
    \label{tbl:distortion}
    
    \fontsize{7pt}{7pt}\selectfont
    \renewcommand{\arraystretch}{1.15}
    \hspace*{-1.32cm}
    \begin{tabular}{|c|cc|cc|cc|cc|}
        \hline
        & \multicolumn{8}{c|}{Method} \\
        \cline{2-9} & \multicolumn{2}{c|}{Uniform Sampling} & \multicolumn{2}{c|}{Lightweight} & \multicolumn{2}{c|}{Welterweight} & \multicolumn{2}{c|}{Fast Coreset} \\
        & Streaming & Static & Streaming & Static & Streaming & Static & Streaming & Static \\
        \cline{2-9}
        $c$-outlier & \underline{\textbf{221 $\pm$ 15K}} & \underline{\textbf{261 $\pm$ 44K}} & 1.07 $\pm$ 0.0 & \textbf{5.51 $\pm$ 78.9} & 1.09 $\pm$ 0.0 & \underline{\textbf{12.1 $\pm$ 80.1}} & 1.13 $\pm$ 0.0 & 1.11 $\pm$ 0.0 \\
        Geometric & \underline{\textbf{66.5 $\pm$ 2.7K}} & \underline{\textbf{140 $\pm$ 1.8K}} & \underline{\textbf{85.2 $\pm$ 2.8K}} & \underline{\textbf{45.6 $\pm$ 4.2K}} & 1.09 $\pm$ 0.0 & \underline{\textbf{10.4 $\pm$ 349}} & 1.15 $\pm$ 0.0 & 1.12 $\pm$ 0.0 \\
        Gaussian Mix. & 1.51 $\pm$ 0.07 & 2.42 $\pm$ 2.52 & 2.35 $\pm$ 0.67 & 2.38 $\pm$ 1.78 & 1.45 $\pm$ 0.05 & 3.65 $\pm$ 3.85 & 1.15 $\pm$ 0.0 & 1.24 $\pm$ 0.0 \\
        Benchmark & 1.10 $\pm$ 0.0 & 1.07 $\pm$ 0.0 & 1.08 $\pm$ 0.0 & 1.11 $\pm$ 0.0 & 1.09 $\pm$ 0.0 & 1.11 $\pm$ 0.0 & 1.18 $\pm$ 0.0 & 1.16 $\pm$ 0.0 \\
        MNIST & 1.42 $\pm$ 0.0 & 1.08 $\pm$ 0.0 & 1.07 $\pm$ 0.0 & 1.07 $\pm$ 0.0 & 1.02 $\pm$ 0.0 & 1.09 $\pm$ 0.0 & 1.12 $\pm$ 0.0 & 1.08 $\pm$ 0.0 \\
        Adult & 1.33 $\pm$ 0.0 & \underline{\textbf{895K $\pm$ 3.2B}} & 1.09 $\pm$ 0.0 & 1.09 $\pm$ 0.0 & 1.27 $\pm$ 0.01 & 1.32 $\pm$ 0.0 & 1.14 $\pm$ 0.0 & 1.15 $\pm$ 0.0 \\
        \hline
    \end{tabular}
    \caption{Distortion means and variances in the streaming and non-streaming setting for $k$-means; taken over 5 runs. Failure cases (distortion $>5$) are bolded. Catastrophic failures (distortion $>10$) are underlined.}
    \label{tbl:composition}
    
    \fontsize{9pt}{9pt}\selectfont
    \renewcommand{\arraystretch}{1.15}
    \begin{tabular}{|l|ccc|}
        \hline
        & \multicolumn{2}{c}{Static} & \multirow{2}{*}{Streaming} \\
        & $m=40k$ & $m=80k$ & \\
        \cline{2-4}
        $c$-outlier & \textbf{\underline{12.5 $\pm$ 0.60}} & \textbf{6.10 $\pm$ 3.90} & \textbf{\underline{18.8 $\pm$ 5.47}} \\
        Geometric & 2.42 $\pm$ 0.01 & 1.38 $\pm$ 0.0 & 2.91 $\pm$ 0.14 \\
        Gaussian Mix. & \textbf{\underline{11.4 $\pm$ 0.69}} & \textbf{7.58 $\pm$ 0.33} & \textbf{\underline{27.0 $\pm$ 6.72}} \\
        Benchmark & \textbf{5.49 $\pm$ 0.03} & 2.64 $\pm$ 0.0 & \textbf{6.67 $\pm$ 0.01} \\
        -- & -- & -- & -- \\
        MNIST & \textbf{6.41 $\pm$ 1.36} & 3.83 $\pm$ 0.02 & \textbf{\underline{11.7 $\pm$ 2.53}} \\
        Adult & 3.12 $\pm$ 0.0 & 3.39 $\pm$ 0.0 & \textbf{9.03 $\pm$ 3.10} \\
        Star & 1.07 $\pm$ 0.0 & 1.26 $\pm$ 0.0 & -- \\
        Song & \textbf{5.44 $\pm$ 1.32} & 3.24 $\pm$ 0.45 & -- \\
        Cover-Type & 1.49 $\pm$ 1.78 & 1.41 $\pm$ 0.0 & -- \\
        Taxi & 1.60 $\pm$ 0.0 & 1.84 $\pm$ 0.0 & -- \\
        Census & 3.45 $\pm$ 0.03 & 3.13 $\pm$ 0.04 & -- \\
        \hline
    \end{tabular}
    \caption{Distortion values for the BICO algorithm in the static and streaming settings, taken over five runs. Failure cases (distortion $>5$) are bolded. Catastrophic failures (distortion $>10$) are underlined.}
    \label{tbl:bico_distortions}
\end{table*}
\renewcommand{\arraystretch}{1}

We use $j$ going forward to describe the number of centers in the candidate $j$-means solution. Thus, lightweight coresets have $j=1$ while Fast-Coresets have
$j=k$.

We take a moment here to motivate the welterweight coreset algorithm.  Consider that lightweight coresets use the $1$-means solution to obtain the sensitivities that dictate the sampling distribution whereas sensitivity sampling uses the full $k$-means solution. In effect, as we change the value of $j$, the cluster sizes $|\calC_p|$ in our approximate solution change. Referring back to equation \ref{eq:sensitivity}, one can see that setting $j$ between $1$ and $k$ acts as an interpolation between uniform and sensitivity sampling.  We default to $j = \log k$ unless stated otherwise. We use the term `accelerated sampling methods' when referring to uniform, lightweight and welterweight coresets as a group.

\paragraph{Datasets.} We complement our suite of real datasets with the following artificial datasets. We default to $n = 50\,000$ and $d=50$ unless stated otherwise.
\begin{description}
    \item \emph{- c-outlier}. Place $n-c$ points in a single location and $c$ points a large distance away.
    \item \emph{- Geometric}. Place $c k$ points at $(1, 0, 0, \cdots)$, $\frac{ck}{r}$ points at $(0, 1, 0, \cdots)$, $\frac{ck}{r^2}$ points
        at $(0, 0, 1, \cdots)$, and so on for $\log_r (ck)$ rounds. Thus, the data creates a high-dimensional simplex with uneven weights across the vertices. We
        default to $c = 100$ and $r=2$.
    \item \emph{- Gaussian mixture}. A set of scattered Gaussian clusters of varying density.

        These clusters are sequentially defined, with the size of the first cluster defined by $\frac{n}{\kappa} \exp \left( \gamma \cdot \rho_0 \right)$, where
        $\kappa$ is the number of Gaussian clusters, $\rho_0$ is uniformly chosen from $[-0.5, 0.5]$, and $\gamma$ is a hyperparameter that affects the
        distribution of cluster sizes.  Then, given clusters $\{c_1, \cdots, c_i\}$, we obtain the size of the $(i+1)$-st cluster by \[ \quad \quad \quad \quad |c_{i+1}|
        = \frac{n - \sum_i |c_i|}{\kappa - i}\exp \left( \gamma \cdot \rho_{i+1} \right).\]

        This has the property that all clusters have size $n / k$ when $\gamma = 0$ and, as $\gamma$ grows, the cluster sizes diverge at an exponential rate.
        We note that this is a well-clusterable instance with respect to cost stability conditions, see \cite{AwS12,Cohen-AddadS17,KuK10,ORSS12}.

    \item \emph{- Benchmark}. A specific distribution of points introduced in \cite{chrisESA} as a testbed for coreset algorithms.  It has the property that all
        reasonable $k$-means solutions are of equal quality but are maximally far apart in the solution space. Thus, the dataset is fully determined by the
        number of centers $k$. As suggested, we produce three benchmark datasets of varying size before applying random offsets to each. We choose the sizes by
        $k_1 = \frac{k}{c_1}$, $k_2 = \frac{k - k_1}{c_2}$, and $k_3 = k - k_1 - k_2$ for $c_1, c_2 \in \mathbb{R}^+$. Note, the benchmark dataset being difficult for sensitivity sampling does not imply that it should be equally difficult for other sampling methods.

\end{description}

The artificial datasets are constructed to emphasize strengths and weaknesses of the various sampling schemas. For example, the $c$-outlier problem contains
very little information and, as such, should be simple for any sampling strategy that builds a reasonable representation of its input. The geometric dataset
then increases the difficulty by having more regions of interest that must be sampled. The Gaussian mixture dataset is
harder still, as it incorporates uneven inter-cluster distances and inconsistent cluster sizes. Lastly, the benchmark dataset is devised to be a worst-case
example for sensitivity sampling.

\paragraph*{Data Parameters}
\label{app:data_params}

In all real and artificial datasets, we add random uniform noise $\eta$ with $0 \leq \eta_i \leq 0.001$ in each dimension in order to make all points unique. Unless specifically varying these parameters, we default all algorithms in~\ref{ssec:algorithms} to $k=100$ for the Adult, MNIST, Star, and artificial datasets and $k=500$ for the Song, Cover Type, Taxi, and Census datasets. Our default coreset size is then $m = 40k$. We refer to the coreset size scalar (the ``$40$'' in the previous sentence) as the \emph{$m$-scalar}.  We only run the dimension-reduction step on the MNIST dataset, as the remaining datasets already have sufficiently low dimensionality.  We run our experiments on an Intel Core i9 10940X 3.3GHz 14-Core processor.

\subsection{Evaluating Sampling Strategies}
\label{ssec:alg_qualities}

\paragraph*{Theoretically guaranteed methods.}

We first round out the comparison between the Fast-Coreset algorithm and standard sensitivity sampling. Specifically, the last columns of Tables~\ref{tbl:distortion} and~\ref{tbl:composition} show that the Fast-Coreset method produces compressions of consistently low distortion and that this holds across datasets, $m$-scalar values and in the streaming setting.  Despite this, Figure~\ref{fig:coreset_size_on_sens_quality} shows that varying $k$ from $50$ to $400$ causes a linear slowdown in sensitivity sampling but only a logarithmic one for the Fast-Coreset method. This analysis confirms the theory in Section~\ref{sec:theory} -- Fast-Coresets obtain equivalent compression to sensitivity sampling but do not have a linear runtime dependence on $k$. We therefore do not add traditional sensitivity sampling to the remaining experiments.

\paragraph*{Speed vs. Accuracy.}

\begin{figure}
\centering
\begin{tabular}{lc}
    \includegraphics[width=\linewidth]{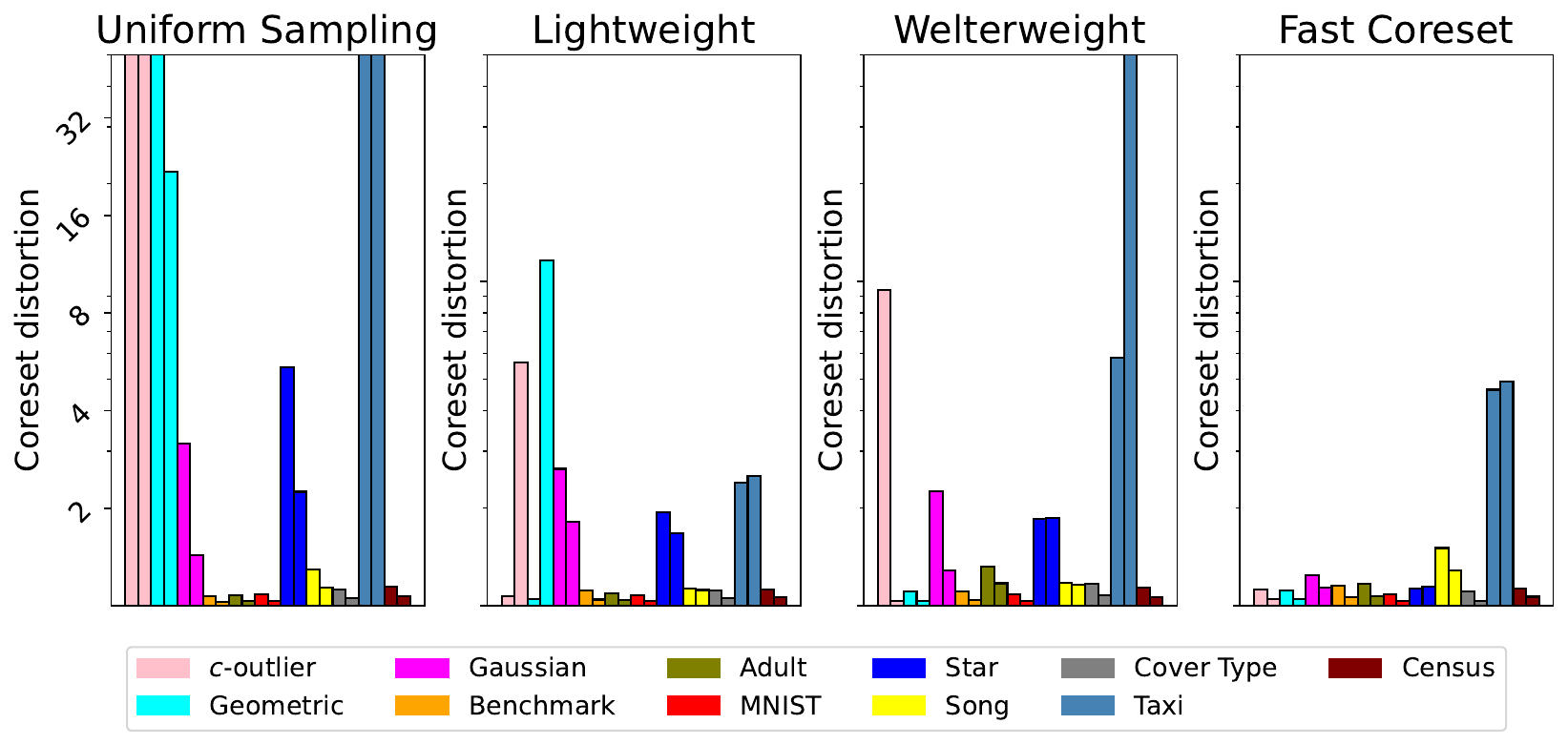} \\
    \includegraphics[width=\linewidth]{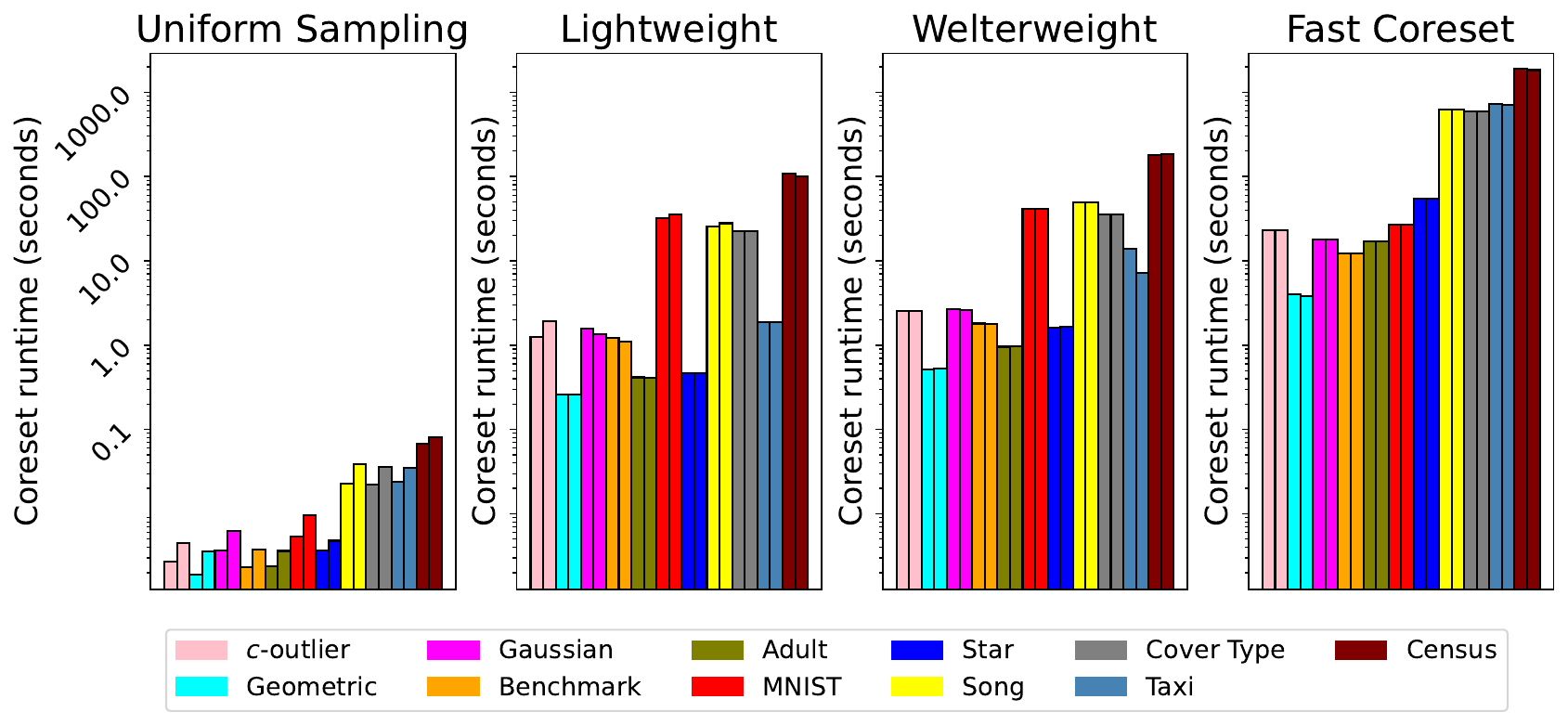}
\end{tabular}

\caption{\emph{Top}: The effect of the $m$-scalar on coreset distortion for real-world datasets. This is a visualization of the data in
Table~\ref{tbl:distortion}.  \emph{Bottom}: The effect of the $m$-scalar on the algorithm runtime for real-world datasets. All values are the mean over 5 runs.
The three bars represent samples of size $m=40k, 80k$.}

\label{fig:coreset_size_on_quality}
\end{figure}

We now refer the reader to the remaining columns of Table~\ref{tbl:distortion} and to Figure~\ref{fig:coreset_size_on_quality}, where we show the effect of coreset size across datasets by sweeping over $m$-scalar values. Despite the suboptimal theoretical guarantees of the accelerated sampling methods, we see that they obtain competitive distortions on most of the real-world datasets while running faster than Fast-Coresets in practice. However, uniform sampling breaks on the Taxi and Star datasets -- Taxi corresponds to the 2D start locations of taxi rides in Porto and has many clusters of varied size while Star is the pixel values of an image of a shooting star (most pixels are black except for a small cluster of white pixels).  Thus, it seems that uniform sampling requires well-behaved datasets, with few outliers and consistent class sizes.

To verify this, consider the results of these sampling strategies on the artificial datasets in Table~\ref{tbl:distortion} and Figure~\ref{fig:coreset_size_on_quality}: as disparity in cluster sizes and distributions grows, the accelerated sampling methods have difficulty capturing all of the outlying points in the dataset. Thus, Figure~\ref{fig:coreset_size_on_quality} shows a clear interplay between runtime and sample quality: \emph{the faster the method, the more brittle its compression}.

\begin{figure}
\centering
\hspace*{-0.1cm}
\includegraphics[trim={2cm 0 2cm 0},clip,width=0.7\linewidth]{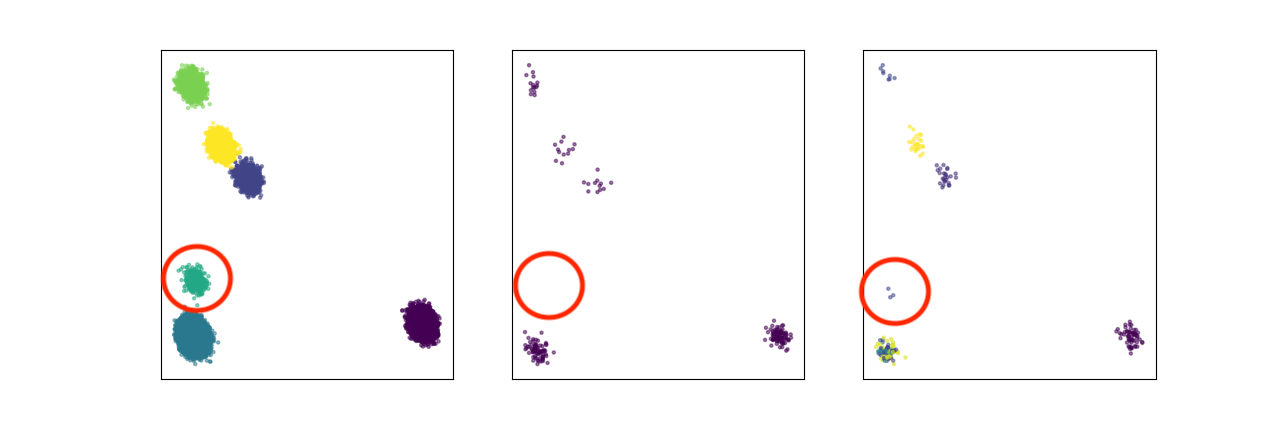}
\vspace*{-0.8cm}
\caption{
The results of lightweight and fast-coreset constructions on a 2D Gaussian mixture dataset of $n=100K$ points with clusters of varying size. The circled cluster has $\sim 400$
points and coresets have 200 points.
\emph{Left}: Original multivariate-Gaussian dataset.
\emph{Middle}: Lightweight coresets fail to capture the cluster of $\sim$ 400 points.
\emph{Right}: Sensitivity sampling with $j=k$ identifies all of the clusters.
}
\label{fig:lightweight_breaks}
\end{figure}

While uniform sampling is expected to be brittle, it may be less obvious what causes light- and welterweight coresets to break. The explanation is simple for lightweight coresets: they sample according to the $1$-means solution and are therefore biased towards points that are far from the mean. Thus, as a simple counterexample, lightweight coresets are likely to miss a small cluster that is close to the center-of-mass of the dataset. This can be seen in Figure~\ref{fig:lightweight_breaks}, where we show an example where the lightweight coreset construction fails on the Gaussian mixture dataset. Since the small circled cluster is close to the center-of-mass of the dataset, it is missed when sampling according to distance from the mean.

Generalizing this reasoning also explains the brittleness of welterweight coresets when $j<k$. To see this, let $\calC_j$ be the approximation obtained during welterweight coresets and observe that the sum of importance values of the points belonging to center $c_i \in \calC_j$ is \[ \sum_{p \in c_i} \left[ \dfrac{\cost(p, c_i)}{\cost(c_i, \calC_j)} + \dfrac{1}{|c_i|} \right] = \dfrac{\sum_{p \in c_i} \cost(p, c_i)}{\cost(c_i, \calC_j)} + 1 = 2.\] Thus, our probability mass is distributed across the clusters that have been found in the approximate solution. Naturally, if $j < k$ and we missed a cluster from $\opt$, there is some set of points that have not received an appropriate probability mass and may therefore be missed.

\vspace*{0.3cm}
\begin{minipage}{0.44\textwidth}
    \fontsize{9pt}{9pt}\selectfont
    \renewcommand{\arraystretch}{1.15}
    \begin{tabular}{|l|cccc|}
        \hline
        & $\gamma = 0$ & $\gamma = 1$ & $\gamma = 3$ & $\gamma = 5$\\
        \hline
        LW Coreset & 1.03 & 1.03 & 1.36 & 2.17\\
        $j=2$ & 1.04 & 1.04 & 1.04 & 1.92\\
        $j=\log k$ & 1.04 & 1.04 & 1.04 & 1.95\\
        $j=\sqrt{k}$ & 1.05 & 1.06 & 1.04 & 1.18\\
        Fast Coreset & 1.03 & 1.03 & 1.04 & 1.12\\
        \hline
    \end{tabular}
    \vspace*{0.1cm}
    \captionof{table}{The effect of $\gamma$ in the Gaussian mixture dataset on the coreset distortion. We report the means over 5 random dataset generations.
    Each generation had 50\,000 points in 50 dimensions, with 50 Gaussian clusters and coresets of size 4\,000. We set $k=100$. }
    \label{tbl:class-imbalance}
\end{minipage}
\quad
\begin{minipage}{0.5\textwidth}
    \fontsize{9pt}{9pt}\selectfont
    \renewcommand{\arraystretch}{1.15}
    \begin{tabular}{|c|cccc|}
        \hline
        & \makecell{Uniform\\Sampling} & \makecell{Light-\\weight} & \makecell{Welter-\\weight} & \makecell{Fast\\Coreset} \\
        \hline
        MNIST & 125 & 124 & 125 & 124 \\
        Adult & 250 & 249 & 249 & 250 \\
        Star & 24.6 & 53.9 & 25.9 & 27.3 \\
        Song & 168 & 166 & 170 & 170 \\
        Census & 367 & 356 & 374 & 322 \\
        Taxi & 1.42 & 0.13 & 0.10 & 0.13\\
        Cov. Type & 105 & 103 & 100 & 98 \\
        \hline
    \end{tabular}
    \vspace*{0.1cm}
    \captionof{table}{$cost(P, \calC_S)$, where $P$ is the whole dataset and $\calC_S$ is found via $k$-means++\cite{ArV07, cohen2020fast} ($k=50$) and Lloyd's algorithm on the coreset. Sample sizes are
    $m=4\,000$ for the first two rows and $m=20\,000$ for the remaining ones. Initializations are identical within each row. We show the first 3 digits of the cost for readability.}
    \label{tbl:lloyds}
\end{minipage}

We evaluate the full extent of this relationship in Table~\ref{tbl:class-imbalance}, where we show the interplay between the welterweight coreset's $j$ parameter (number of centers in the approximate solution) and the Gaussian mixture dataset's $\gamma$ parameter (higher $\gamma$ leads to higher class imbalance). We can consider this as answering the question ``How good must our approximate solution be before sensitivity sampling can handle class imbalance?'' To this end, all the methods have low distortion for small values of $\gamma$ but, as $\gamma$ grows, only Fast-Coresets (and, to a lesser extent, welterweight coresets for larger values of $j$) are guaranteed to have low distortion.

\begin{figure}
    \centering
    \includegraphics[width=\linewidth]{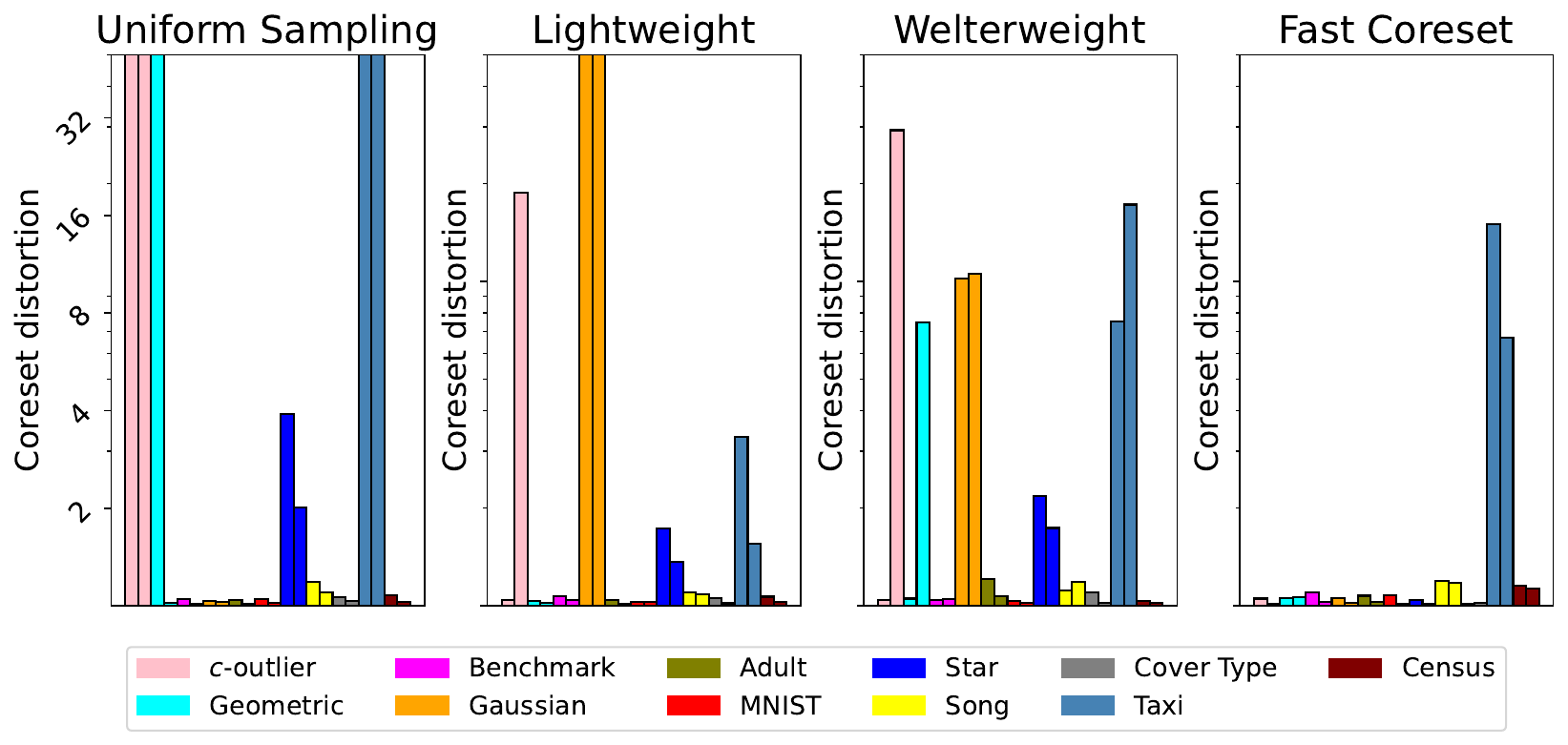}
    \vspace*{-0.6cm}
    \caption{Sample coreset distortions under $k$-median for one run on each dataset. Bars within each dataset correspond to $m=40k, 60k, 80k$.}
    \label{subfig:kmedian_distortion}
\end{figure}
For completeness, we verify that these results also hold for the $k$-median task in Figure~\ref{subfig:kmedian_distortion}. There, we see that $k$-median distortions across datasets are consistent with $k$-means distortions. We show one of five runs to emphasize the random nature of compression quality when using various sampling schemas.

To round out the dataset analysis, we note that BICO performs consistently poorly on the coreset distortion metric\footnote{We do not include BICO or Streamkm++ in Figures \ref{fig:coreset_size_on_quality}, \ref{subfig:kmedian_distortion}, \ref{fig:streaming_runtimes},  as they do not fall into the $\tilde{O}(nd)$ complexity class and are only designed for $k$-means.}, as can be seen in Table \ref{tbl:bico_distortions}. We also analyze the Streamkm++ method across the artificial datasets in Table~\ref{tbl:streamkm++} with $m=40k$ and see that it obtains poor distortions compared to sensitivity sampling. This is due to Streamkm++'s required coreset size -- logarithmic in $n$ and exponential in $d$ -- being much larger than those for sensitivity sampling (sensitivity sampling coresets depend on neither parameter). We did not include Streamkm++ in tables~\ref{tbl:distortion},~\ref{tbl:composition} due to its suboptimal coreset size, distortion and runtime.

Lastly, we point out that every sampling method performs well on the benchmark dataset, which is designed to explicitly punish sensitivity sampling's reliance on the initial solution. Thus, we verify that there is no setting that breaks sensitivity sampling.

\begin{table}
    \centering
    \begin{tabular}{lcccc}
        \hline
        Dataset & $c$-outlier & Geometric & Gaussian & Benchmark \\
        Distortion & 2.07 & 1.40 & 1.54 & 2.53 \\
        \hline
    \end{tabular}
    \caption{Distortions for Streamkm++ on artificial datasets.}
    \label{tbl:streamkm++}
\end{table}

We lastly show how well these compression schemas facilitate fast clustering on large datasets in Table~\ref{tbl:lloyds}. Consider that a large coreset-distortion means that the centers obtained on the coreset poorly represent the full dataset. However, among sampling methods with small distortion, it may be the case that one consistently leads to the `best' solutions. To test this, we compare the solution quality across all fast methods on the real-world datasets, where coreset distortions are consistent. Indeed, Table~\ref{tbl:lloyds} shows that no sampling method leads to solutions with consistently minimal costs.

\subsection{Streaming Setting}
\label{ssec:streaming}

\begin{figure}
    \centering
    \hspace*{0.1cm}\includegraphics[width=0.985\linewidth]{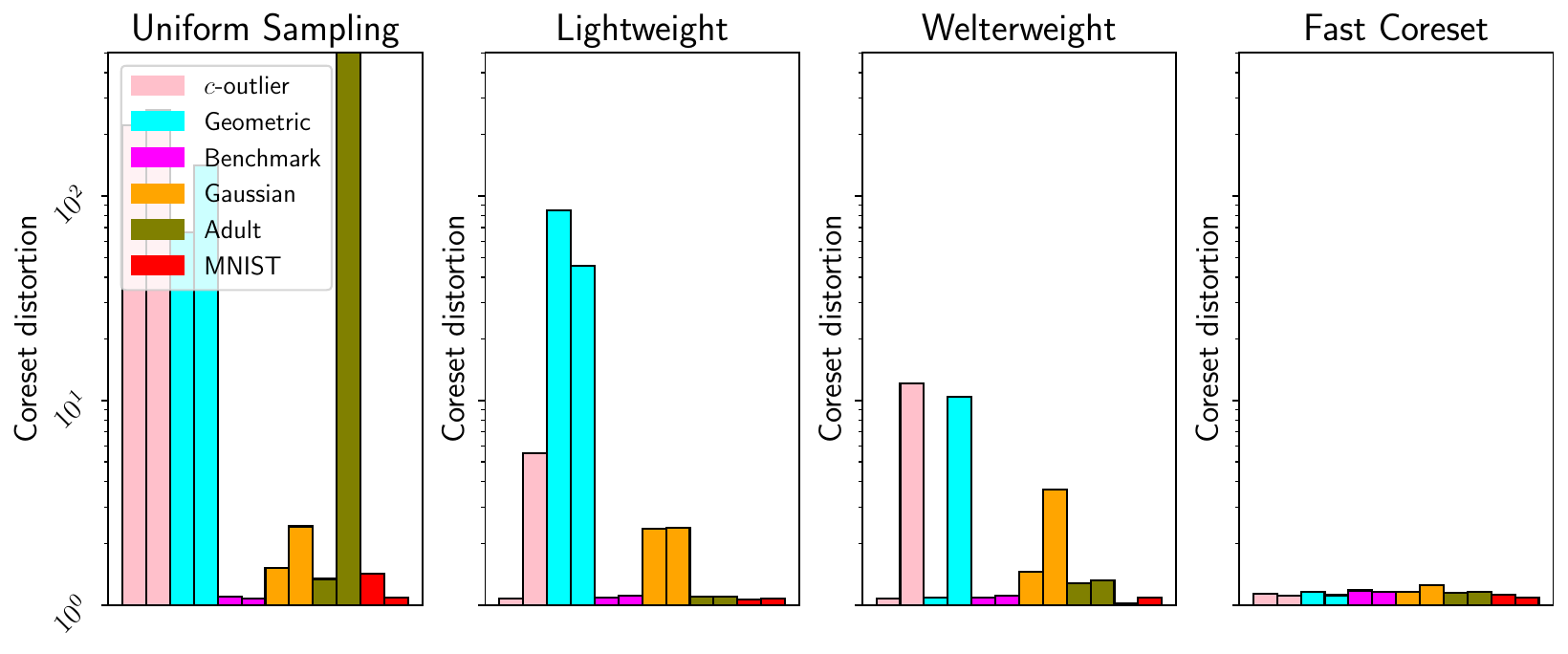}
    \includegraphics[width=\linewidth]{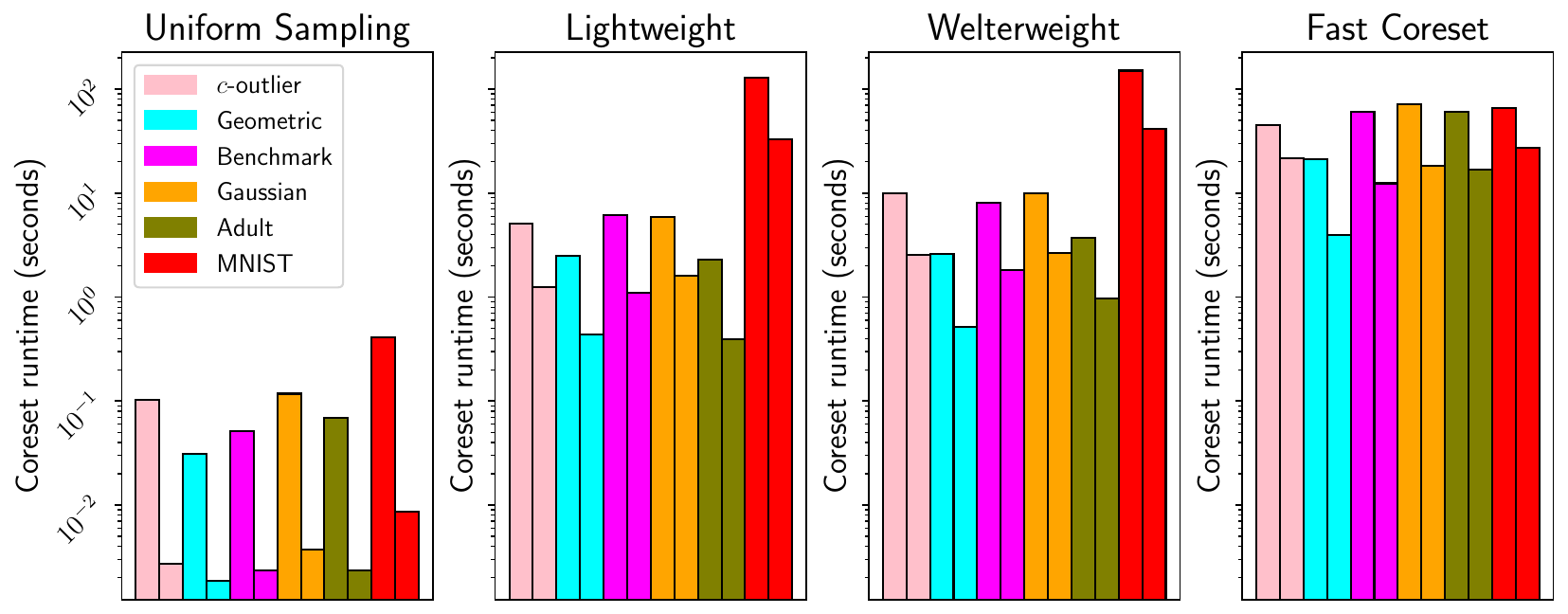}

    \caption{\emph{Top}: Coreset distortion on the $k$-means task in the streaming and non-streaming settings. This is a visualization of the data in
    Table~\ref{tbl:composition}. \emph{Bottom}: Coreset construction runtimes in the streaming and non-streaming settings for the linear and sub-linear
    complexity coreset algorithms. Bars are [Streaming, Non-Streaming].}

    \label{fig:streaming_runtimes}
\end{figure}

One of the most common use-cases for big-data algorithms is the streaming setting, where one receives input in batches and must maintain a compression that is representative of the dataset. Although there is a wealth of sampling and coreset methods in the streaming paradigm, we require consistency across algorithms and therefore assume a black-box sampling procedure. Since the coreset property is preserved under composition, we utilize the merge-\&-reduce strategy originally proposed by \cite{BS80} and first applied to maintaining clustering coresets in stream by \cite{HaM04}. The idea is to first partition the input into $b$ blocks and then perform sampling and composition along them until a single compression is obtained. Specifically, we start by obtaining a coreset on each block. Then, combining samples using a complete binary tree, we (1) recursively re-sample from the children until there is at least one coreset for each level\footnote{If there are $b=8$ blocks, then there would be coresets corresponding to blocks $[[1], [2], [3, 4], [5, 6, 7, 8]]$} in the tree and then (2) concatenate these samples and obtain one final coreset from the composition. Since we are composing coresets from coresets, the errors stack and, in theory, we should require more samples to obtain a similar accuracy.

Despite this, we see the opposite result for many of our sampling strategies. Surprisingly, Table~\ref{tbl:composition} and Figure~\ref{fig:streaming_runtimes} show that the accelerated sampling methods all perform \emph{at least as well} under composition on the artificial datasets and do not suffer significant drops in accuracy, variance or runtime on the real datasets. Although inconsistent with the prevailing intuition, we must therefore conclude that the accelerated sampling methods are \emph{equally} feasible in the streaming setting.  We suspect that this is due to the non-uniformity imposed by the merge-\&-reduce algorithm. To see this, consider uniform sampling on the $c$-outlier dataset during the final step of the composition, where we are composing the samples corresponding to each layer of the tree. Assume first that our outlier points happened to fall in the first block. Then we have taken a sample of size $m$ from this block and immediately use this for the final composition. Thus, in this case the outliers are \emph{more} likely to be in our final sample than in the non-streaming setting. In the alternate setting where the outliers are less likely, our expected error is already high and missing the outlier `more' cannot worsen our expected error. 

\subsection{Takeaways}
To summarize the comparisons between sampling strategies and datasets, the practical guideline is that uniform sampling usually works well, so an optimistic user can default to that (while accepting that it might fail). This was evidenced in Sections~\ref{ssec:intro_experiments} and \ref{ssec:alg_qualities}. However, it was also shown there that the accelerated sampling methods all have a chance of catastrophically failing. Thus, in accordance with Table~\ref{tbl:class-imbalance}, a cautious user may wish to verify whether uniform sampling will work by checking how balanced the dataset's clusters are. By our theoretical contribution (Corollary~\ref{cor:mainAlg} and Theorem~\ref{thm:main_theorem}), performing this verification is just as expensive as computing a coreset. Thus, the cautious user may as well create a Fast-Coreset in that amount of time. Importantly, we do not claim that there is a `best' algorithm among the ones that have been discussed.

\section{Conclusion}

In this work, we discussed the theoretical and practical limits of compression algorithms for center-based clustering. We proposed the first nearly-linear time
coreset algorithm for $k$-median and $k$-means. Moreover, the algorithm can be parameterized to achieve an asymptotically optimal coreset size. Subsequently, we
conducted a thorough experimental analysis comparing this algorithm with fast sampling heuristics. In doing so, we find that although the Fast-Coreset algorithm
achieves the best compression guarantees among its competitors, naive uniform sampling is already a sufficient compression for downstream clustering tasks in
well-behaved datasets. Furthermore, we find that intermediate heuristics interpolating between uniform sampling and coresets play an important role in
balancing efficiency and accuracy. 

Although this closes the door on the highly-studied problem of optimally small and fast coresets for $k$-median and $k$-means, open questions of wider scope
still remain. For example, when does sensitivity sampling guarantee accurate compression with optimal space in linear time and can these
conditions be formalized? Furthermore, sensitivity sampling is incompatible with paradigms such as fair-clustering
\cite{BandyapadhyayFS21, BravermanCJKST022, cohen2019fixed, HJV19, SSS19} and it is unclear whether one can expect that a linear-time method can optimally
compress a dataset while adhering to the fairness constraints.

\section{Acknowledgements}

Andrew Draganov and Chris Schwiegelshohn are partially supported by the Independent Research Fund Denmark (DFF) under a Sapere Aude Research Leader grant No 1051-00106B. David Sauplic has received funding from the European Union’s Horizon 2020
research and innovation programme under the Marie Sklodowska-Curie grant
agreement No 101034413.

\section{Proofs, Pseudo-Code, and Extensions}
\label{app:theory}

We put the proofs, pseudo-code and algorithmic extensions towards the end of the paper for improved readability of the primary text.  Algorithm
\ref{alg:crudeApx} corresponds to the discussion in Section~\ref{ssec:crude_bound} and Algorithm~\ref{alg:reduce-diam} corresponds to the discussion in
Section~\ref{ssec:reduce_spread}.

\label{app:pseudoCode}
\begin{algorithm}[tb]
   \caption{Crude-Approx($P$)}
   \label{alg:crudeApx}
\begin{algorithmic}[1]
\Procedure{Count-Distinct-Cells}{$P, c, \ell$} \Comment data $P$, $c$ is the center of the quadtree grid at level $\ell$
   \State Let $\calD$ be a dictionary, and count $= 0$.
   \For{each point $p \in P$}
   \State let $c^p$ be the center of the cell containing $p$. The $i$-th coordinate of $c_p$ is $\lfloor \frac{p_i - c_i}{2^\ell}\rfloor  \cdot 2^\ell + \frac{2^{\ell}}{2}$. 
   \If{$c_p$ is a not a key of $\calD$} 
   \State insert $c_p$ in $\calD$ and do count $\gets$ count $+1$.
   \EndIf
   \EndFor
   \State {\bfseries Output:} True if count $\geq k+1$, False otherwise.
   \EndProcedure
   
   \Procedure{Crude-Approx}{$P$} \Comment data $P$, with diameter $\Delta$.
   \State let $s$ be a u.a.r in $[0, \Delta]^d$, and $c = (s,..., s) \in \R^d$.
   \State using a binary search, find the smallest $\ell$ such that Count-Distinct-Cells($P, c, \ell$) = True. Let $\ell_0$ be that level.
   \State \textbf{Output:} $U = 2^{\ell_0} \Delta$.
   \EndProcedure
\end{algorithmic}
\end{algorithm}

\begin{algorithm}[tb]
   \caption{Reduce-Spread($P, U$)}
   \label{alg:reduce-diam}
\begin{algorithmic}[1]
\Procedure{Reduce-Diameter}{$P, U$} \Comment data $P$, upper bound on $\opt$ $U$
   \State Let $\calD$ be a dictionary
   \State Let $r = \sqrt d n^2\cdot U$, $s$ be chosen u.a.r. in $\{0, ..., r\}$, and $c = {s, ..., s}$ be the center of the grid.
   \For{each point $p \in P$}
   \State let $c^p$ be the center of the cell containing $p$. The $i$-th coordinate of $c_p$ is $\lfloor \frac{p_i - c_i}{r}\rfloor  \cdot r + \frac{r}{2}$.
   Add $p$ to $\calD[c_p]$.
   \EndFor
   \State Identify the non-empty dictionary keys $c^1, ..., c^k$, and let $\calC_1, ..., \calC_k$, be the corresponding cells.
   \For {each coordinate $i = 1, ..., d$}
   \State Sort the cells according to the $i$-th coordinate of their center. Let $\delta = 0$. 
   \For{$j = 1, ..., k$}
    \If {$\frac{c^j_i - c^j_{i-1}}{r} \geq 2$} update $\delta \gets \delta + c^j_i - c^j_{i-1} - 2r$.
	\EndIf    
    \State Substract $\delta$ from the $i$-th coordinate of all points in the $j$-th cell. 
    \EndFor
   \EndFor
   \State {\bfseries Output:} the dataset $P'$ consisting of all shifted points.
   \EndProcedure
   
   \Procedure{Reduce-Min-Distance}{$P, U$} \Comment data $P$, $U$ such that $U \leq n^2 d \log(\Delta) \opt$.
   \State Round each coordinates of points to the closest multiple of $g := \frac{U}{n^4 d^2 \log \Delta}$.
   \State \textbf{Output:} the dataset $P'$ with all point after rounding.
   \EndProcedure
\end{algorithmic}
\end{algorithm}

\subsection{Proof of Corollary~\ref{cor:mainAlg}}
\label{ssec:cor_proof}
Recall that Corollary~\ref{cor:mainAlg} states that \cref{alg:main} produces an $\eps$-coreset in time $\tilde O(nd \log \Delta)$.
\begin{proof}[Proof of \cref{cor:mainAlg}]
First, performing the Johnson Lindenstrauss embedding \cite{MakarychevMR19} takes time $\tilde O(nd)$.

On the projected dataset, the algorithm \fkmeans runs in time $\tilde O\lpar n \log \Delta\rpar$, and its solution has an approximation-ratio $O\lpar
\tilde{d}^z \log k\rpar = O\lpar\log^{z+1} k\rpar$ for $\tilde P$.  The guarantee offered by the embedding ensures that the clustering $\lbra
\calC_1,...,\calC_k\rbra$ still has approximation ratio for $P$ \cite{MakarychevMR19}. 

For $k$-means, computing the $1$-mean solution for each $\calC_i$ takes time $O(nd)$ (the $1$-mean is simply the mean). 
For $k$-median, computing the $1$-median solution can be done as well in time $O(nd)$ \cite{CohenLMPS16}. 
We note that both may be approximated to a factor $2$ in constant time, by sampling uniformly at randm few points from each cluster \cite{neurips21}.

Provided the $c_i$ and the partition $\calC_i$, computing $|\calC_i|$ and $\cost(\calC_i, c_i)$ for all $i$ also takes time $O(nd)$.

Since the solution consisting of assigning each $p \in \calC_i$ to $c_i$ is a $O\lpar \log^{z+1} k\rpar$-approximation, the values $s(p)$ defined in
\cref{alg:main} can be used to perform the coreset construction algorithm, and we conclude from \cref{fact:logApprox}.

\end{proof}

\subsection{Reduction of $k$-means to $k$-median.}
\label{app:redKM}

In our argument, the only step specific to $k$-median is computing the upper-bound $U$ on the cost of the solution. Provided such an upper-bound, rounding
points and shifting the box would work exactly alike for $k$-means. Therefore, the next lemma is enough to extend our reduction of the spread to $k$-means:

\begin{lemma}\label{lem:kmedTokmeans}
Let $\calS$ be a $c$-approximation for $k$-median on $P$. Then, $\calS$ is a $nc^2$-approximation for $k$-means on $P$.
\end{lemma}

\begin{proof}

Let $\cost_1$ (resp. $\cost_2$) be the $k$-median (resp. $k$-means) cost, $\opt_1$ (resp. $\opt_2$) be the optimal $k$-median (resp. $k$-means) solution. We
have the following inequalities:

\begin{align*}
\cost_2(\calS) &= \sum_{p\in P} \dist(p, \calS)^2 \leq \lpar \sum_{p\in P} \dist(p, \calS)\rpar^2\\
&\leq c^2 \cdot \lpar  \sum_{p\in P} \dist(p, \opt_1)\rpar^2\\
&\leq c^2 \cdot \lpar  \sum_{p\in P} \dist(p, \opt_2)\rpar^2\\
&\leq c^2 \cdot n \cdot  \sum_{p\in P} \dist(p, \opt_2)^2,
\end{align*}
where the last inequality stems from Cauchy-Schwarz. Therefore, $\calS$ is an $nc^2$-approximation to $k$-means. 
\end{proof}

\subsection{Estimation of the Optimal Cost in a Tree}
\label{app:apx-tree-proof}

\begin{proof}[proof of \cref{lem:apxTree}]

If the input is spread in $k+1$ distinct cell at level $i$, then in any solution there is at least one of those cells with no center. In the tree metric, points
lying in this cell have therefore connection cost at least $\sqrt{d}2^{-i+1} \cdot \Delta$. Thus, the left-hand-side of the inequality holds.

On the other hand, if the input is contained in $k$ cells at level $i-1$, then placing a center arbitrarily in each cell yields a solution with cost at most
$n \cdot \sqrt{d}2^{-i+4} \cdot \Delta$. Indeed, the distance from any point to its closest center is at most $2 \cdot \sum_{j \geq i-1} \sqrt{d}2^{-j} \cdot
2\Delta \leq 4 \sqrt{d}2^{-i+2} \cdot \Delta$ (summing the edge length from the point to the cell at level $i$, and then going down to the center). This
concludes the proof.

\end{proof}

\begin{proof}[Proof of \cref{lem:crudeApx}]

The running time of \textsc{Count-Distinct-Cells} in \cref{alg:crudeApx} is $\tilde O(nd)$, using a standard dictionary data structure. To compute $U$,
\textsc{Crude-Approx} makes $O(\log \log \Delta)$ calls to \textsc{Count-Distinct-Cells}, as there are $O(\log \Delta)$ levels. This concludes the complexity of
the algorithm.

The approximation guarantee for $k$-median directly stems from \cref{lem:apxTree} and \cref{lem:quadtreeDist}. For $k$-means, it is a consequence of
\cref{lem:kmedTokmeans}.

\end{proof}

\subsection{Extensions to Algorithm \ref{alg:main}}
\label{app:extensions}

Consider that Algorithm \ref{alg:main} only needs to be provided with an assignment to a solution that is a $O(\polylog k)$ approximation, implying that one
could compute the solution $\calC$ via any algorithm that satisfies this.  Indeed, this initial solution may as well be an $O\lpar\polylog \lnor P \rnor_0 \rpar$
approximation, where $\lnor P \rnor_0$ is the number of non-zero samples in $P$. Using the iterative coreset construction from \cite{BravermanJKW21}, one could
then derive a near-optimal coreset size, only suffering a $O(\log^* n)$ loss in the running time.

As an example, we illustrate a different approach for $k$-median. One could first embed the input into a hierarchically separated tree (HST) with expected
distortion $O(\log \lnor P \rnor_0)$ \cite{FakcharoenpholRT03}. On such tree metrics, solving $k$-median can be done in linear time using dedicated algorithms
(see e.g. \cite{Cohen-AddadLNSS21}). Using the solution from the HST metric, one can compute a coreset, and iterate using the previous argument.  This embedding
into HST is very similar to what is done by the \fkmeans algorithm, but can be actually performed in \emph{any} metric space, not only Euclidean.  For instance,
in a metric described by a graph with $m$ edges, the running time of this construction would be near linear-time $\tilde O(m)$.

\bibliographystyle{plainnat}
\bibliography{references}

\end{document}